\documentclass[a4paper,notitlepage,12pt]{article}
\usepackage[tbtags]{amsmath}
\usepackage{amsfonts}
\usepackage{amsthm}
\usepackage{color}
\usepackage{graphicx,amssymb}
\usepackage{multirow}
\usepackage{rotating}
\usepackage{mathrsfs}
\usepackage{epstopdf}
\usepackage{enumerate}
\usepackage{pdflscape}
\usepackage{geometry}
\usepackage{natbib}
\usepackage{changepage}
\usepackage{dcolumn}
\usepackage{tabularx}
\usepackage{booktabs}
\usepackage{subfig}
\usepackage{mathabx}
\usepackage[mathscr]{euscript}
\usepackage{diagbox}

\usepackage{floatrow}
\usepackage{tikz}
\usetikzlibrary{graphs}
\usepackage{mathdots}
\usepackage{algorithm}
\usepackage{algorithmicx}
\usepackage{algpseudocode}
\newcolumntype{d}[1]{D{.}{.}{#1}}
\newcolumntype{Y}{>{\raggedleft\arraybackslash}X}
\newcolumntype{Z}{>{\centering\arraybackslash}X}

\newtheorem{proposition}{Proposition}

\theoremstyle{remark}

\renewcommand{\arraystretch}{0.85}

\numberwithin{equation}{section}

\usepackage[colorlinks,linkcolor=blue,citecolor=blue]{hyperref}

\title{\vspace{-10mm}When and Where to Transfer: Learning Transferable Historical Contexts for Short-History Multivariate Time Series Forecasting}
\author{
	\textbf{
		Huiling Yuan$^{*1}$,
		Yize Wang \thanks{Equal contribution}$^{*2}$,
		Guodong Li \thanks{Correspondence to gdli@hku.hk} $^{2}$
	} \\[0.6em]
	$^{1}$School of Mathematics and Statistics,
	Donghua University\\
	$^{2}$ Department of Statistics \& Actuarial Science,
	The University of Hong Kong 
}

\begin{document}
\maketitle

\begin{abstract}
Related sources can improve multivariate time series forecasting when the target history is short, but their usefulness varies across source groups and historical periods.
A key challenge is to identify informative historical contexts from the sources.
This paper proposes Hierarchical Context Transfer (HiCT),
a framework that adaptively weights source groups and their historical contexts to determine both \emph{where} and \emph{when} to transfer.
A shared context encoder and cross-attention are used to learn weights over source groups and historical contexts. These weights guide regularized transfer estimation, followed by an optional target-specific correction.
This paper establishes a target excess-risk bound for the context-weighted transfer estimator. Experiments demonstrate competitive predictive performance across four multivariate forecasting datasets and a favorable accuracy--runtime trade-off against the evaluated baselines. Source-corruption ablations illustrate the roles of group-level and within-group weighting in adapting to unreliable source histories.
\end{abstract}

\newpage

\section{Introduction}
\label{sec:introduction}
{Multivariate time series forecasting often involves a target group with a short historical record
	and related source groups with substantially longer histories. Limited target observations make it difficult to estimate temporal and cross-variable dependencies reliably.
	Transfer learning offers a way to incorporate source information into target model estimation \mbox{\citep{li2022translasso,ma2025transfer}}.
	Its benefit depends on whether the transferred information matches the target's current dynamics.}

{This match can vary across source groups and across historical periods within each group. A source group may share predictive structure with the target yet contain periods that are poorly suited to the current forecast. A single weight for an entire source group cannot capture these differences in usefulness. This motivates two related decisions: how much to borrow from each source group and which historical contexts within that group to emphasize. Both decisions should depend on the current target context.}

{With limited target observations, we need a criterion that connects source contributions to target prediction. Historical similarity can guide source selection \mbox{\citep{teller2026volatility}}, but does not directly measure whether a source context improves target prediction after model fitting.
	Validation-based weighting learns sample contributions from held-out prediction loss \mbox{\citep{ren2018reweight}}. For temporal adaptation,
	AdaRNN partitions historical data into periods and matches the distributions of RNN hidden-state representations across them to address temporal covariate shift \mbox{\citep{du2021adarnn}}, while DDG-DA learns resampling probabilities over historical observations to adapt training data to predicted distribution changes \mbox{\citep{li2022ddgda}}.
	Building on prediction-based weighting, we use target prediction loss to learn source contributions across groups and historical contexts.}

We propose \emph{Hierarchical Context Transfer} (HiCT),
a framework that learns \emph{where} and \emph{when} to
transfer for each target forecast.
Conditioned on the current target context, group-level
weights specify how much to borrow from each source group and
within-group weights distribute this contribution across
the group's historical contexts.
Their product gives each source context's weight in the
forecasting loss.
Aligned low-dimensional representations and a shared context
encoder enable transferability scoring across groups with
different observation dimensions.

Target validation loss guides the learning of a shared
weighting function and hyperparameter selection. Given the current
weights, we fit the predictor using target observations and
weighted source contexts, then differentiate validation loss
through this fit to update the weighting parameters.
At test time, the weighting parameters remain fixed.
In the rolling-refit implementation, the function generates
source weights conditioned on each new target context.
These weights guide regularized transfer estimation,
followed by an optional target-specific correction using
the available target history to address residual
source--target mismatch.

{Our contributions are summarized below.}
\begin{itemize}
	\item
	We propose HiCT for multivariate forecasting with short
	target histories. It learns \emph{where} and \emph{when} to transfer through target-conditioned weights over sources and their historical contexts.
	
	\item
	We develop a two-stage estimator that combines regularized
	transfer estimation with target-specific correction.
	We derive a target excess-risk bound for Stage~1 and show
	that exact Stage~2 correction does not increase empirical
	target loss.
	
	\item
	We conduct experiments on four multivariate forecasting
	datasets, showing that HiCT offers strong predictive performance
	at modest computational cost.
	Source-corruption ablations further illustrate the respective
	contributions of group-level and within-group weighting.
\end{itemize}
\section{Related Work}
\label{sec:related_work}

\paragraph{{Selective statistical transfer.}}
{Trans-Lasso combines sparse transfer estimation with a target-specific correction~\mbox{\citep{li2022translasso}}.
	{\citet{park2026lowrank}} select informative sources for low-rank multivariate regression.
	RIW-TL assigns observation-level weights using residual-based density ratios~\mbox{\citep{zhao2026riwtl}}.
	\mbox{\citet{ma2025transfer}} develop transfer estimators for high-dimensional vector autoregressive models. Their method combines low-rank and sparse components and selects informative auxiliary observations.}

\paragraph{{Representation learning and time-series transfer.}}
Target-PCA uses auxiliary panels to improve target factor estimation under missing observations and weak signals~\mbox{\citep{duan2024targetpca}}.
\mbox{\citet{zhou2023limited}} treat historical batches as multiple source domains.
They align source and target feature distributions for forecasting with limited target data. DTr-CNN selects a single source using dynamic time warping
and Jensen--Shannon divergence and transfers a causal CNN
across datasets, with competitive performance on financial
forecasting tasks~\citep{ye2021implementing}. ROSE combines a Transformer architecture with decomposed frequency learning, and register tokens to learn from multiple pre-training domains and adapt to downstream tasks~\mbox{\citep{wang2025rose}}.
These methods focus on transfer across datasets or domains,
without explicitly learning target-conditioned weights over
historical contexts within each source group to guide
forecasting-model estimation.

\paragraph{{Historical selection and retrieval for forecasting.}}
{\mbox{\citet{teller2026volatility}} select source subsequences using dynamic time warping distances.
	These subsequences are combined with target data to train realized-volatility forecasting models.
	RAFT retrieves historical windows and uses their observed continuations to augment predictions~\mbox{\citep{han2025raft}}.
	TimeRAF learns a retriever for a forecasting foundation model~\mbox{\citep{zhang2025timeraf}}.
	SARAF uses stationarity information to regulate retrieval diversity and aggregation
	~\mbox{\citep{zhou2026saraf}}.
	\mbox{\citet{choi2026histories}} learn to rerank historical candidates using future-compatibility supervision during training.}

\paragraph{Validation-based weighting and temporal adaptation.}
\mbox{\citet{ren2018reweight}} obtain sample weights by differentiating validation loss through a training update. Meta-Weight-Net learns a mapping from individual training loss to sample weight using a small unbiased meta-dataset~\mbox{\citep{shu2019metaweightnet}}.
RAM-Net incorporates market regimes into meta-learned sample weighting~\mbox{\citep{jang2025ramnet}}.
AdaRNN weights hidden-state distribution matching across periods~\mbox{\citep{du2021adarnn}}, whereas HiCT's weights
determine each source context's contribution to the forecasting loss.
DDG-DA learns historical-sample resampling probabilities through bilevel optimization with a weighted forecasting proxy~\mbox{\citep{li2022ddgda}}.
HiCT instead learns a shared function from one-step-ahead target validation loss to assign source-group and historical-context weights for each target context.

\section{Learning Transferable Historical Contexts}
\label{section3}

\begin{figure}[h]
	\centering
	\includegraphics[width=\linewidth]{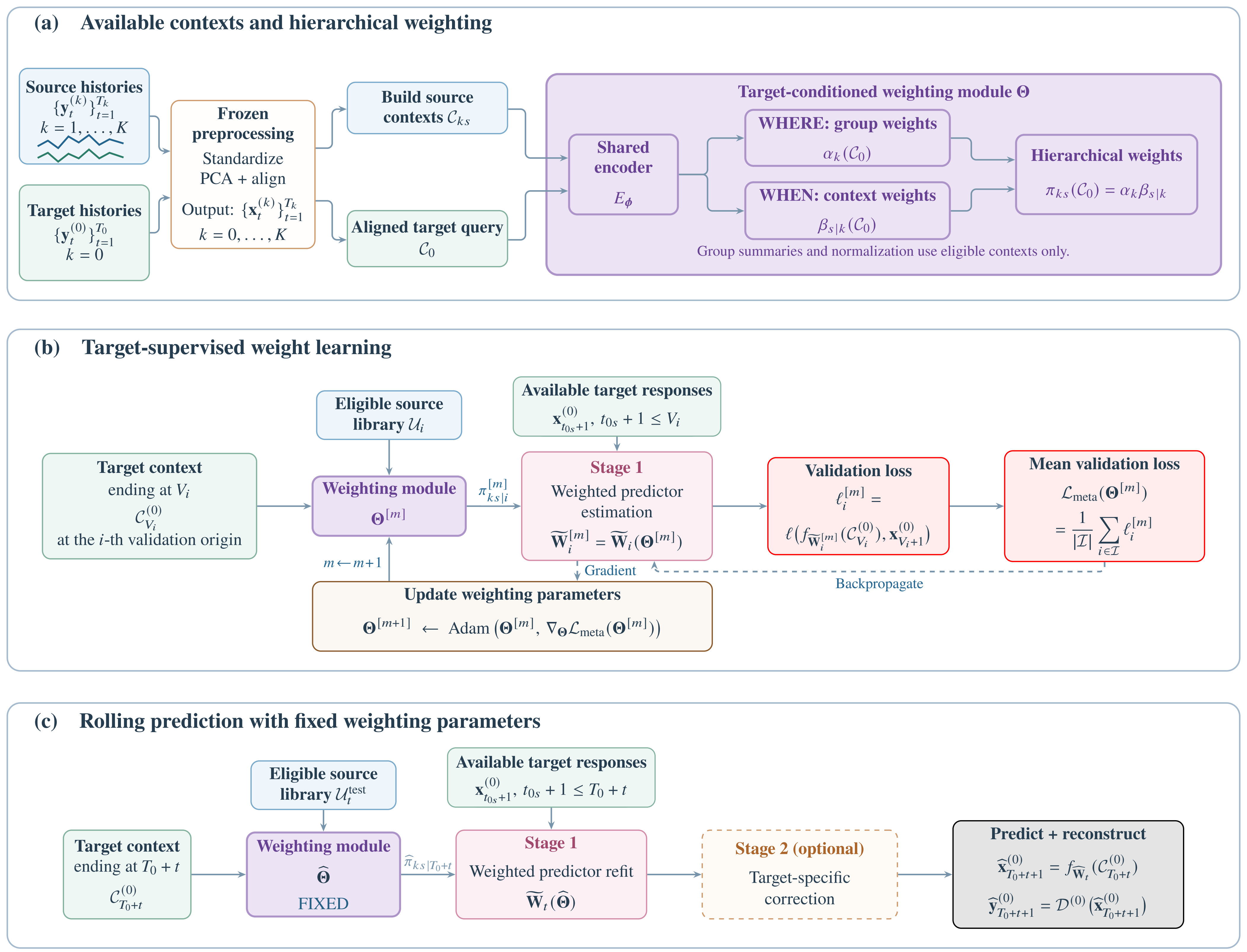}
	\caption{Overview of HiCT. (a) Aligned contexts receive target-conditioned group and within-group weights. (b) Weighting parameters are learned by differentiating the mean loss over all target validation queries through weighted predictor fits. (c) With these parameters fixed, each test context generates new weights for predictor refitting, optional target-specific correction, and one-step-ahead forecast reconstruction.}
	\label{hict:overview}
\end{figure}

HiCT learns target-conditioned weights over source groups and their historical contexts as showed in Figure~\mbox{\ref{hict:overview}}. The parameters $\boldsymbol\Theta$ define the weight-generating function, whose outputs $\pi_{ks}$ depend on the target context; $\mathbf W$ contains the prediction parameters in the weighted forecasting loss. Learning alternates predictor fitting with validation-based updates of $\boldsymbol\Theta$. At test time, $\widehat{\boldsymbol\Theta}$ is fixed, while the rolling-refit procedure regenerates weights and fits prediction parameters for each target context, with an optional target-specific correction. 

\subsection{Problem Setup and Group-Time Contexts}
\label{section3.1}

We consider one target group, indexed by $k=0$, and $K$ source groups, indexed by $k=1,\ldots,K$.
{For each group $k$, let $N_k$ denote its observation dimension.
	The observations from group $k$ form a multivariate time series.}
\begin{align*}
	\mathbf{y}_t^{(k)}
	&=
	\left(
	y_{1t}^{(k)},\ldots,y_{N_k t}^{(k)}
	\right)^\top
	\in \mathbb{R}^{N_k},
	\qquad
	t=1,\ldots,T_k,
\end{align*}
{where $T_k$ is the number of observed time points in group $k$ and $y_{it}^{(k)}$ denotes the value of variable $i$ at time $t$. The superscript $(k)$ identifies the group, with $(0)$ denoting the target group. We use $t$ as a within-group observation index and write $\tau_t^{(k)}$ for its calendar time.}
We focus on the short-history setting, where the target history is substantially shorter than every source history.
Specifically, we assume $T_0 \ll T_k$ for all $k=1,\ldots,K$.

Since the group dimensions $N_k$ may differ, we map each multivariate series to an aligned $d$-dimensional representation $\mathbf{x}_t^{(k)} \in \mathbb{R}^d$, as described in Section~\ref{section3.2}. Given the observed target history, we aim to forecast $\mathbf{y}_{T_0+1}^{(0)}$ by exploiting transferable information from the source groups.
{A single weight for each source group cannot distinguish the transferability of different historical contexts within that group.} We therefore model transferability at both the group and temporal levels to learn \emph{where} and \emph{when} to transfer.

\paragraph{Historical contexts.}

To represent variation in transferability over time,
we construct rolling historical contexts of length $L$
with step size $h$.
We take $L$ and $h$ to be positive integers and assume
$T_k\geq L+1$ for all groups.
For each group $k=0,\ldots,K$, the context ending at
an observed time $t\geq L$ is
\begin{equation*}
	\mathcal C_t^{(k)}
	=
	\left(
	\mathbf x_{t-L+1}^{(k)},\ldots,
	\mathbf x_t^{(k)}
	\right)^\top
	\in\mathbb R^{L\times d}.
\end{equation*}
We sample historical contexts at $t_{ks}=L+(s-1)h$,
$s=1,\ldots, S_k$, where $S_k=1+\left\lfloor{(T_k-L-1)}/{h}\right\rfloor$ is the number of context--response pairs in group $k$.
For brevity, we write
$\mathcal C_{ks}=\mathcal C_{t_{ks}}^{(k)}$.
The index $s$ identifies the sampled context,
and $t_{ks}$ is its ending time.
Since $t_{ks}+1\leq T_k$, each context $\mathcal C_{ks}$
has an observed one-step-ahead response
$\mathbf x_{t_{ks}+1}^{(k)}$.
For source groups $k=1,\ldots,K$, this context--response pair forms a group--time transfer unit indexed by $(k,s)$, with $k$ and $s$ specifying \emph{where} and \emph{when}
to transfer from, respectively.
At the current forecasting origin $T_0$, we write
$\mathcal C_0=\mathcal C_{T_0}^{(0)}$
for the target query context used to forecast
$\mathbf x_{T_0+1}^{(0)}$.

In Figure~\ref{hict:overview}(b), $V_i$ denotes the forecasting-origin index of validation query $i$, whose context is $\mathcal C_{V_i}^{(0)}$. Section~\ref{section3.4} defines the validation schedule.
In panel~(c), rolling test prediction starts at $T_0$.
The target context at test step $t=0,1,\ldots$
is $\mathcal C_{T_0+t}^{(0)}$.
We write $\mathcal U_t^{\mathrm{test}}$ for the library
of source contexts whose one-step-ahead responses are
available at this forecasting origin.
As new target observations become available,
we extend the same sampling grid and include historical
target samples whose responses satisfy
$t_{0s}+1\leq T_0+t$.

\subsection{Hierarchical Transferability Weighting}
\label{section3.2}
To score contexts across groups with different numbers of observed variables, we first construct aligned $d$-dimensional representations.
A shared Transformer encoder then maps the aligned contexts to features used by the group-level and within-group context weighting branches.

\textbf{Aligned representations.}
For each group $k=0,\ldots,K$, the map $\mathcal P^{(k)}:\mathbb R^{N_k}\to\mathbb R^d$ yields latent observations $\mathbf z_t^{(k)}=\mathcal P^{(k)}(\mathbf y_t^{(k)})$. These maps include preprocessing and are fitted using training data; our experiments use principal component analysis (PCA) as in Appendix~\ref{app:preprocessing}. A common dimension does not ensure comparable coordinates, so we align each source with the target. Let $\mathcal M_k^{\mathrm{com}}$ contain training-index pairs $(u,v)$ with $\tau_u^{(0)}=\tau_v^{(k)}$. We estimate an orthogonal matrix in $\mathbb R^{d\times d}$ for each source by
\begin{equation*}
	\widehat{\mathbf A}^{(k)}
	=\operatorname*{arg\,min}_{\mathbf A^\top\mathbf A=\mathbf I_d}
	\sum_{(u,v)\in\mathcal M_k^{\mathrm{com}}}
	\left\|\mathbf z_u^{(0)}-\mathbf A\mathbf z_v^{(k)}\right\|_2^2,
	\qquad k=1,\ldots,K,
\end{equation*}
where $\mathbf I_d$ is the $d\times d$ identity matrix. The aligned observations are $\mathbf x_t^{(0)}=\mathbf z_t^{(0)}$ and $\mathbf x_t^{(k)}=\widehat{\mathbf A}^{(k)}\mathbf z_t^{(k)}$ for $k\geq1$. This construction assumes that latent coordinates are approximately comparable up to rotation and reflection.

\textbf{Shared contextual encoding.}
A shared Transformer $E_{\boldsymbol\phi}$ maps the aligned
target and source contexts to
$\mathbf H_0=E_{\boldsymbol\phi}(\mathcal C_0)$ and
$\mathbf H_{ks}=E_{\boldsymbol\phi}(\mathcal C_{ks})$
in $\mathbb R^{L\times d_h}$, where $d_h$ is the hidden dimension and the vector $\boldsymbol\phi$ collects the shared encoder parameters.
The encoder uses an input projection, learnable positional
embeddings, multi-head self-attention, and position-wise
feed-forward blocks.
Cross-attention then summarizes each source context using
target embeddings as queries and source embeddings
as keys and values:
\begin{equation}
	\operatorname{CrossAttn}_{\boldsymbol\vartheta}
	(\mathbf H_0,\mathbf H_{ks})
	=
	\operatorname{Pool}_t
	\left[
	\operatorname{softmax}
	\left(
	\frac{
		(\mathbf H_0\mathbf W_Q)
		(\mathbf H_{ks}\mathbf W_K)^\top
	}{
		\sqrt{d_a}
	}
	\right)
	\mathbf H_{ks}\mathbf W_V
	\right]
	\in\mathbb R^{d_a}.
	\label{eq:cross-attention}
\end{equation}
Here,
$\boldsymbol\vartheta
=\{\mathbf W_Q,\mathbf W_K,\mathbf W_V\}$
contains learnable projection matrices in
$\mathbb R^{d_h\times d_a}$, where $d_a$ is the attention dimension.
The softmax normalizes over source time steps, and
$\operatorname{Pool}_t$ averages over target time steps.
The group-level and context-level branches use separate
attention parameters, included in $\boldsymbol\phi_g$
and $\boldsymbol\phi_c$, respectively.
These two parameter collections also contain the corresponding scoring-function parameters; the roman superscripts $\mathrm g$ and $\mathrm c$ label the group and context branches, respectively.

{\textbf{Group-level transferability: where to transfer.}
	For each source group $k$, we average its encoded historical
	contexts to obtain a group-level representation,}
\begin{equation*}
	\text{{$\displaystyle \mathbf H_k^{\mathrm g}
			=
			\frac{1}{S_k}
			\sum_{s=1}^{S_k}\mathbf H_{ks}
			\in\mathbb R^{L\times d_h}.$}}
\end{equation*}
{We use \eqref{eq:cross-attention} and a scoring function
	to obtain a target-conditioned group summary and its
	relevance score,}
\begin{equation*}
	\text{{$\displaystyle \mathbf z_k^{\mathrm g}
			=
			\operatorname{CrossAttn}_{\boldsymbol\phi_g}
			\left(\mathbf H_0,\mathbf H_k^{\mathrm g}\right),
			\qquad
			a_k
			=
			g_{\boldsymbol\phi_g}^{\mathrm g}
			\left(\mathbf z_k^{\mathrm g}\right),$}}
\end{equation*}
{where $\mathbf z_k^{\mathrm g}\in\mathbb R^{d_a}$ and
	$g_{\boldsymbol\phi_g}^{\mathrm g}:\mathbb R^{d_a}\to\mathbb R$
	is a learnable scoring function.
	Normalizing these scores across source groups yields}
\begin{equation*}
	\text{{$\displaystyle \alpha_k(\mathcal C_0)
			=
			\frac{\exp(a_k)}{\sum_{j=1}^{K}\exp(a_j)},
			\qquad k=1,\ldots,K.$}}
\end{equation*}
{The weight $\alpha_k(\mathcal C_0)$ determines
	\emph{where} to transfer by specifying how much to borrow
	from source group $k$.}

\textbf{Context-level transferability: when to transfer.}
Similarly, we compute a target-conditioned summary and
relevance score for each source context,
\begin{equation*}
	\text{{$\displaystyle \mathbf z_{ks}^{\mathrm c}
			=
			\operatorname{CrossAttn}_{\boldsymbol\phi_c}
			\left(\mathbf H_0,\mathbf H_{ks}\right),
			\qquad
			b_{ks}
			=
			g_{\boldsymbol\phi_c}^{\mathrm c}
			\left(\mathbf z_{ks}^{\mathrm c}\right),$}}
\end{equation*}
{where $\mathbf z_{ks}^{\mathrm c}\in\mathbb R^{d_a}$ and
	$g_{\boldsymbol\phi_c}^{\mathrm c}:\mathbb R^{d_a}\to\mathbb R$
	is a learnable scoring function.
	Normalizing these scores within group $k$ yields}
\begin{equation*}
	\text{{$\displaystyle \beta_{s\mid k}(\mathcal C_0)
			=
			\frac{\exp(b_{ks})}{\sum_{r=1}^{S_k}\exp(b_{kr})},
			\qquad s=1,\ldots,S_k.$}}
\end{equation*}
{These weights determine \emph{when} to transfer by specifying
	which historical contexts to emphasize within each source group.}

{\textbf{Joint transferability weights.}
	Combining the group-level and context-level weights gives}
\begin{equation*}
	\text{{$\displaystyle \pi_{ks}(\mathcal C_0)
			=
			\alpha_k(\mathcal C_0)\beta_{s\mid k}(\mathcal C_0).$}}
\end{equation*}
{These weights define a normalized distribution over
	source contexts,}
\begin{equation}
	\text{{$\displaystyle \pi_{ks}(\mathcal C_0)\geq 0,
			\qquad
			\sum_{k=1}^{K}\sum_{s=1}^{S_k}
			\pi_{ks}(\mathcal C_0)=1.$}}
	\label{normalized_weights}
\end{equation}
{Write $\boldsymbol\pi=(\pi_{ks})$ for the joint weight vector. For brevity, we omit the dependence on $\mathcal C_0$ when it is clear from context.}

\subsection{Context-Weighted Transfer and Target Adaptation}
\label{section3.3}

For given transferability weights, we estimate the forecasting
parameters in two stages.
Let $f_{\mathbf W}$ map an aligned historical context to
its one-step-ahead prediction in the aligned latent space.
For predictor parameters $\mathbf W$ and a prediction loss $\ell$, we
define the source-context and empirical target losses as
\begin{equation*}
	\begin{gathered}
		{\displaystyle L_{ks}(\mathbf W)=\ell\!\left(f_{\mathbf W}(\mathcal C_{ks}),\mathbf x_{t_{ks}+1}^{(k)}\right),} \quad
		{\displaystyle L_0(\mathbf W)=\frac{1}{S_0}\sum_{i=1}^{S_0}\ell\!\left(f_{\mathbf W}(\mathcal C_{0i}),\mathbf x_{t_{0i}+1}^{(0)}\right),}
	\end{gathered}
\end{equation*}
{where $\mathcal C_{0i}$ is the $i$-th historical target context and $S_0\geq1$ is the number of available target forecasting samples.}
We assume that the two optimization problems below admit minimizers.

\textbf{Stage 1: Context-weighted transfer estimation.}
For the current target context $\mathcal C_0$, let
$\pi_{ks}=\widehat\pi_{ks}$ denote the weights from
\eqref{eq:final-transferability}, with zero weights assigned
to unavailable source contexts.
The transferred estimate is
\begin{equation}
	\widetilde{\mathbf W}
	\in
	\arg\min_{\mathbf W}
	\left\{
	L_0(\mathbf W)
	+
	\lambda_{\mathrm{tr}}
	\sum_{k=1}^{K}\sum_{s=1}^{S_k}
	\pi_{ks}L_{ks}(\mathbf W)
	+
	\lambda_{\mathbf W}\mathcal J(\mathbf W)
	\right\},
	\label{eq:weighted-transfer}
\end{equation}
where $\lambda_{\mathrm{tr}}\geq 0$ controls the contribution
of source contexts and $\lambda_{\mathbf W}\geq 0$ sets
the strength of the model-specific regularizer $\mathcal J$,
such as an $\ell_1$ (lasso) or nuclear-norm penalty.

\textbf{Stage 2: Target-specific adaptation.}
To address residual source--target mismatch, we fix
$\widetilde{\mathbf W}$ and estimate a correction
$\boldsymbol\Delta$ using only the target history,
\begin{equation}
	\widehat{\boldsymbol\Delta}
	\in
	\arg\min_{\boldsymbol\Delta}
	\left\{
	L_0\left(\widetilde{\mathbf W}+\boldsymbol\Delta\right)
	+
	\lambda_\Delta\mathcal J_\Delta(\boldsymbol\Delta)
	\right\},
	\label{eq:target-debiasing}
\end{equation}
where $\lambda_\Delta\geq 0$ controls the regularization
strength and $\mathcal J_\Delta$ encodes structural
assumptions on the source--target discrepancy.
The final estimator is
$\widehat{\mathbf W}
=\widetilde{\mathbf W}+\widehat{\boldsymbol\Delta}$.
With $\widehat{\boldsymbol\Delta}=\mathbf 0$ when
target-specific correction is omitted, the one-step-ahead latent forecast and its target-space reconstruction are
\begin{equation}
	{\displaystyle \widehat{\mathbf x}_{T_0+1}^{(0)}=f_{\widehat{\mathbf W}}(\mathcal C_0),
		\qquad
		\widehat{\mathbf y}_{T_0+1}^{(0)}=\mathcal D^{(0)}\!\left(\widehat{\mathbf x}_{T_0+1}^{(0)}\right),}
	\label{eq:forecast-reconstruction}
\end{equation}
{where $\mathcal D^{(0)}:\mathbb R^d\to\mathbb R^{N_0}$ is the reconstruction map associated with the target representation. Appendix~\mbox{\ref{app:preprocessing}} describes reconstruction and evaluation.}

Proposition~\ref{prop:1} establishes a target excess-risk bound for Stage~1 under the stated generalization-gap bound. Proposition~\ref{prop:2} guarantees non-increase of the empirical target loss under exact Stage~2 correction.
Appendix~\ref{app:theory} provides the proof of
Proposition~\ref{prop:1} and the statement and proof of Proposition~\ref{prop:2}.

Let $\mathcal E$ be the finite-dimensional prediction-parameter space,
and let $\mathcal W\subseteq\mathcal E$ denote the parameter class.
The population risks are defined as
$R_0(\mathbf W)=\mathbb E[L_0(\mathbf W)]$ and
$R_{ks}(\mathbf W)=\mathbb E[L_{ks}(\mathbf W)]$.
Let $\mathbf W_0^*$ be a minimizer of $R_0$ over $\mathcal W$.
Define
$\delta_{ks}=\sup_{\mathbf W\in\mathcal W}
|R_{ks}(\mathbf W)-R_0(\mathbf W)|$.
The functions $L_{\boldsymbol\pi}$ and $R_{\boldsymbol\pi}$
denote the target-plus-weighted-source empirical loss and
population risk, respectively, both normalized by $1+\lambda_{\mathrm{tr}}$.

\begin{proposition}[Stage 1: Risk bound for context-weighted transfer estimation]
	\label{prop:1}
	Under the assumptions in Appendix~\ref{app:transfer-theory},
	let the transferability
	weights $\boldsymbol\pi$ be fixed and satisfy
	\eqref{normalized_weights}, with zero weights assigned
	to unavailable source contexts.
	Suppose that
	\begin{equation}
		\sup_{\mathbf W\in\mathcal W}
		\left|
		L_{\boldsymbol\pi}(\mathbf W)
		-
		R_{\boldsymbol\pi}(\mathbf W)
		\right|
		\leq
		\varepsilon_{\mathrm{gen}}(\boldsymbol\pi),
		\label{eq:weighted-risk-deviation}
	\end{equation}
	for some finite $\varepsilon_{\mathrm{gen}}(\boldsymbol\pi)\geq 0$.
	Let $\varepsilon_{\mathrm{opt}}\geq0$ denote the normalized
	objective tolerance in \eqref{eq:approximate-transfer-estimation}. Then any transferred estimator
	$\widetilde{\mathbf W}\in\mathcal W$ satisfying
	\eqref{eq:approximate-transfer-estimation} obeys
	\begin{equation}
		\begin{aligned}
			R_0(\widetilde{\mathbf W})-R_0(\mathbf W_0^*)
			\leq {}&
			\frac{2\lambda_{\mathrm{tr}}}{1+\lambda_{\mathrm{tr}}}
			\sum_{k=1}^{K}\sum_{s=1}^{S_k}
			\pi_{ks}\delta_{ks}
			+\frac{\lambda_{\mathbf W}}{1+\lambda_{\mathrm{tr}}}
			\mathcal J(\mathbf W_0^*)
			+2\varepsilon_{\mathrm{gen}}(\boldsymbol\pi)
			+\varepsilon_{\mathrm{opt}}.
		\end{aligned}
		\label{prop:context-transfer}
	\end{equation}
\end{proposition}

Concentrating all weight on the source context with the
smallest $\delta_{ks}$ minimizes the discrepancy term in
the risk bound, but can increase estimation variance.
Appendix~\ref{app:transfer-theory} illustrates this
bias--variance trade-off through a simplified
weighted-mean model.

\subsection{Target-Supervised Learning of Transferability Weights}
\label{section3.4}

We learn the weighting parameters $\boldsymbol\Theta=(\boldsymbol\phi,\boldsymbol\phi_g,\boldsymbol\phi_c)$ from the last $T_{\mathrm{val}}$ target observations, indexed by the validation set $\mathcal I=\{1,\ldots,T_{\mathrm{val}}\}$. For each $i\in\mathcal I$, set $V_i=T_0-T_{\mathrm{val}}+i-1$ and hold out $\mathbf x_{V_i+1}^{(0)}$ as the response. We assume $1\leq T_{\mathrm{val}}\leq T_0-L-1$ so that each query has a preceding target context--response pair for fitting. Each weighting-parameter update uses the full validation set,
with source eligibility at origin $V_i$ determined by
response availability in calendar time:
\begin{equation}
	\mathcal U_i
	=\left\{(k,s):
	1\leq k\leq K,\;1\leq s\leq S_k,\;
	\tau_{t_{ks}+1}^{(k)}\leq\tau_{V_i}^{(0)}
	\right\}.
	\label{eq:available-source-library}
\end{equation}
The module assigns nonnegative weights $\pi_{ks\mid i}(\boldsymbol\Theta)$ summing to one over $\mathcal U_i$. Group summaries and both weighting branches use only eligible contexts; groups with no eligible contexts receive zero weight. Let $\mathcal D_{<i}$ contain the target pairs $(\mathcal C_{0s},\mathbf x_{t_{0s}+1}^{(0)})$ with $t_{0s}+1\leq V_i$. The held-out response is excluded from its own predictor fit and the weighting module's direct inputs, while earlier validation
responses may enter later fits once observed.

Let $\widetilde{\mathbf W}_i(\boldsymbol\Theta)$ minimize
\eqref{eq:weighted-transfer} with the target
pairs in $\mathcal D_{<i}$ and the
weighted source contexts in $\mathcal U_i$. The validation objective evaluates these predictors before target-specific correction:
\begin{equation}
	\mathcal L_{\mathrm{meta}}(\boldsymbol\Theta)
	=\frac{1}{|\mathcal I|}\sum_{i\in\mathcal I}
	\ell\left(
	f_{\widetilde{\mathbf W}_i(\boldsymbol\Theta)}(\mathcal C_{V_i}^{(0)}),
	\mathbf x_{V_i+1}^{(0)}
	\right).
	\label{eq:meta-transfer-objective}
\end{equation}
We update $\boldsymbol\Theta$ with Adam by differentiating
through the weighted predictor fits.
For iterative fitting, each
$\widetilde{\mathbf W}_i(\boldsymbol\Theta)$ is approximated
by $M\geq1$ differentiable prediction-parameter updates.
Appendix~\ref{app:weight-optimization} provides the
optimization details and final weight construction.

In the rolling-refit implementation, we fix $\widehat{\boldsymbol\Theta}$ and apply \eqref{eq:available-source-library} at $T_0+t$ to obtain $\mathcal U_t^{\mathrm{test}}$. Each current target context generates weights normalized over this library. We refit the predictor using these weights and available observations, then forecast after optional target-specific correction with Algorithm~\ref{alg:hict} in appendix ~\ref{app:A}.

\section{Experiments}
\label{section:4}

We evaluate forecasting performance on Stock, NY Taxi, PEMS-BAY, and Electricity, then compare computational cost before examining the weighting components through source-corruption ablations and sensitivity analysis. Dataset construction and preprocessing are detailed in Appendix~\ref{app:datasets}.

\subsection{Forecasting Performance and Computational Cost}
\label{section:4.1}

For each dataset, we split the target history chronologically into training, validation, and test periods. Preprocessing transformations are fitted on training data and held fixed thereafter.
Hyperparameters are selected using validation data.
We evaluate HiCT with a linear vector autoregressive
predictor (HiCT-VAR) and recurrent predictors (HiCT-RNN).

We compare HiCT with three target-only baselines and two
transfer methods that incorporate source data.
Among the target-only baselines,
PatchTST~\citep{nie2023time} uses a patch-based Transformer,
TimeMixer~\citep{wang2024timemixer} mixes temporal patterns
across scales with MLPs, and AdaRNN~\citep{du2021adarnn}
performs temporal adaptation within the target history
through recurrent networks.
The two source-based transfer methods are DTr-CNN and ROSE.
DTr-CNN~\citep{ye2021implementing} transfers a CNN from
a single source to the target.
For ROSE~\citep{wang2025rose}, we pre-train a Transformer
on the available source groups and fine-tune it on target data.
Appendix~\ref{app:baseline-implementations} lists details the adaptations made to each
baseline for our evaluation protocol.
We report mean squared error (MSE) and mean absolute error
(MAE) in the standardized target test set.

Table~\ref{tab:main-results} reports forecasting performance
on the four datasets, and Table~\ref{tab:runtime-stock}
compares total fitting and forecasting time on Stock.
Overall, HiCT achieves the lowest MSE and MAE on three of
the four datasets, with no consistent advantage from
replacing the linear predictor with a recurrent model.
HiCT-VAR performs best on PEMS-BAY and Electricity, while
HiCT-RNN leads on NY Taxi.
DTr-CNN achieves the lowest errors on Stock but requires
more computation time than HiCT-VAR, which ranks second
in both metrics.
The Transformer-based PatchTST and ROSE are substantially
more time-consuming than HiCT, whose Transformer is used
only for the weighting function rather than forecasting.
Appendix~\ref{app:runtime} describes the timing protocol
and predictor updates.

\begin{table}[H]
	\centering
	\normalsize
	\setlength{\tabcolsep}{4pt}
	\renewcommand{\arraystretch}{1.1}
	\caption{One-step-ahead errors in the full standardized target space. Bold and underlining mark the lowest and second-lowest distinct errors, respectively; ties share markings.}
	\label{tab:main-results}
	\begin{tabular}{lrrrrrrrr}
		\toprule
		& \multicolumn{2}{c}{Stock} & \multicolumn{2}{c}{NY Taxi}
		& \multicolumn{2}{c}{PEMS-BAY} & \multicolumn{2}{c}{Electricity} \\
		\cmidrule(lr){2-3} \cmidrule(lr){4-5} \cmidrule(lr){6-7} \cmidrule(lr){8-9}
		Method & MSE & MAE & MSE & MAE & MSE & MAE & MSE & MAE \\
		\midrule
		PatchTST & 1.6798 & 0.9902 & 0.4226 & 0.4464 & 0.1073 & 0.1966 & 0.1192 & 0.2242 \\
		TimeMixer & 1.2410 & 0.8291 & 0.3655 & 0.4301 & 0.1301 & 0.2264 & 0.1363 & 0.2424 \\
		AdaRNN & 1.0412 & 0.7654 & 0.3216 & 0.3987 & 0.2970 & 0.3446 & 0.3360 & 0.3592 \\
		DTr-CNN & \textbf{0.8254} & \textbf{0.6276} & 0.3285 & 0.4001 & 0.1087 & 0.1844 & 0.1916 & 0.2841 \\
		ROSE & 1.1456 & 0.7835 & 0.3119 & 0.3857 & 0.1081 & 0.1870 & {0.1045} & \underline{0.2142} \\
		HiCT-VAR & \underline{0.8448} & \underline{0.6649} & \underline{0.2764} & \underline{0.3558} & \textbf{0.0993} & \textbf{0.1749} & \textbf{0.0944} & \textbf{0.2107} \\
		HiCT-RNN & 0.9183 & 0.7065 & \textbf{0.2687} & \textbf{0.3487} & \underline{0.1064} & \underline{0.1827} & \underline{0.0989} & 0.2148 \\
		\bottomrule
	\end{tabular}
\end{table}

\begin{table}[H]
	\centering
	\normalsize
	\setlength{\tabcolsep}{4pt}
	\renewcommand{\arraystretch}{1.1}
	\caption{Total fitting and forecasting time on Stock (eight CPU threads), including any rolling refits. }
	\label{tab:runtime-stock}
	\begin{tabular}{lrrrrrrr}
		\toprule
		Method & PatchTST & TimeMixer & AdaRNN & DTr-CNN & ROSE & HiCT-VAR & HiCT-RNN \\
		\midrule
		Total (s) & 2942.6 & 29.5 & 4.3 & 56.9 & 527.5 & 9.3 & {10.9} \\
		\bottomrule
	\end{tabular}
\end{table}

\subsection{{Ablation and Sensitivity Analyses}}
\label{section:4.2}

We examine group-level and within-group context weighting
through two source-corruption experiments on U.S. Stock.
The first replaces the entire available Industrials input
sequence with independent zero-mean Gaussian noise.
The second applies the same corruption to the later
chronological half of each source group's available input
sequence. Noise is scaled by the corresponding training-coordinate standard deviations, while target observations remain unchanged. We refer to the original condition as \emph{unperturbed sources}.

For each source condition, we learn the weighting function
using target validation loss.
We then remove one weighting component while keeping the
fitted scorer and predictor settings fixed.
The w/o learned $\alpha$ variant sets $\alpha_k=1/4$ for
all four source groups and retains learned $\beta$.
The w/o learned $\beta$ variant uses uniform within-group
weights and retains learned $\alpha$.
Target-only excludes the source loss while preserving the
target representation and predictor penalty.
These comparisons assess adaptation through weight learning
under source corruption, rather than the response of a
frozen model trained on unperturbed sources.

\begin{table}[H]
	\centering
	\normalsize
	\setlength{\tabcolsep}{4pt}
	\renewcommand{\arraystretch}{1.1}
	\caption{Forecasting errors under source corruption on Stock. Bold indicates the lowest error for each metric within each condition. Parentheses report percentage reductions relative to Target-only, calculated from the displayed errors.
		Negative values indicate deterioration.}
	\label{tab:ablation-main}
	\begin{tabular}{llrr}
		\toprule
		Source condition & Method & MSE & MAE \\
		\midrule
		\multirow{2}{*}{Unperturbed} & Target-only & 0.9818 (0.00\%) & 0.7308 (0.00\%) \\
		& HiCT & \textbf{0.8457} (13.86\%) & \textbf{0.6662} (8.84\%) \\
		\midrule
		\multirow{2}{*}{\shortstack[l]{Industrials corrupted\\(entire history)}} & w/o $\alpha$ & {0.8782} (10.55\%) & {0.6824} (6.62\%) \\
		& HiCT & \textbf{0.8696} (11.43\%) & \textbf{0.6789} (7.10\%) \\
		\midrule
		\multirow{2}{*}{\shortstack[l]{All sources corrupted\\(later half)}} & w/o $\beta$ & 0.9900 (-0.84\%) & 0.7151 (2.15\%) \\
		& HiCT & \textbf{0.9719} (1.01\%) & \textbf{0.7086} (3.04\%) \\
		\bottomrule
	\end{tabular}
\end{table}

\begin{figure}[H]
	\centering
	\includegraphics[width=\linewidth]{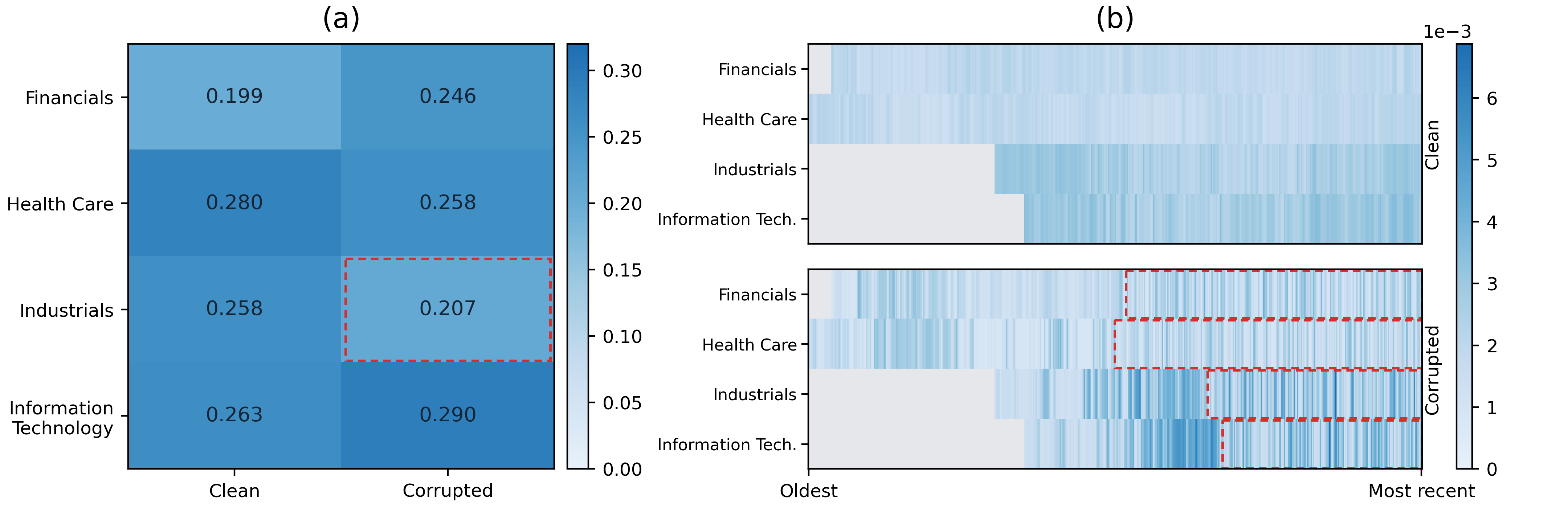}
	\caption{Mean transferability weights on Stock:
		(a) group weights $\alpha$ when the entire Industrials input
		history is corrupted; (b) conditional within-group weights
		$\beta$ when the later half of each source's input history
		is corrupted. Weights are averaged over test forecasting origins.
		Red boxes mark the corrupted group or contexts; gray denotes
		padding. Each panel uses a common color scale across conditions.}
	\label{fig:corruption-weights}
\end{figure}

When the entire Industrials input history is replaced by
noise, HiCT achieves lower MSE and MAE than the equal-group
weighting variant in Table~\ref{tab:ablation-main}.
The lower mean weight assigned to Industrials in
Figure~\ref{fig:corruption-weights}(a) is consistent with
reduced reliance on the corrupted source.
These results support learned group weighting in this setting.

When the later half of each source's input history is
corrupted, HiCT also achieves lower errors than uniform
within-group weighting.
Both errors remain below Target-only, whereas removing
learned $\beta$ raises MSE above this baseline.
Figure~\ref{fig:corruption-weights}(b) shows increased weights
on some uncorrupted contexts in Industrials and Information
Technology. The uncorrupted half also receives greater total weight
in front half of Industrials and Information Technology.

\begin{figure}[H]
	\centering
	\includegraphics[width=\linewidth]{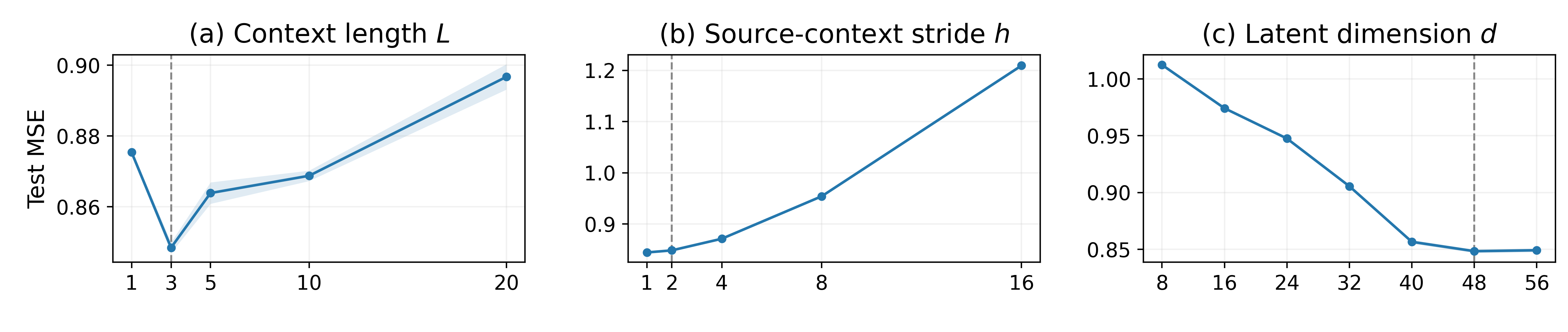}
	\caption{Test sensitivity on Stock. Each sweep varies one parameter,
		fixing the others at $(L,h,d)=(3,2,48)$. Curves and shading show
		the mean and one sample standard deviation over five runs;
		dashed lines mark validation-selected values.}
	\label{fig:stock-sensitivity}
\end{figure}

Figure~\ref{fig:stock-sensitivity} examines sensitivity to context length
$L$, source-context stride $h$, and latent dimension $d$ on Stock.
Extending the context beyond the selected short window or using
coarser source sampling increases test error, while gains from
increasing $d$ level off.
Appendix~\ref{app:sensitivity} provides the corresponding validation
curves and discusses parameter selection.
Appendix~\ref{app:weight-visualization} presents heatmaps
of the learned transferability weights across source
groups and historical contexts.

\section{Conclusion}
\label{sec:conclusion}

We proposed HiCT for multivariate forecasting with short target histories.
By learning \emph{where} and \emph{when} to transfer, HiCT draws on
relevant information across source groups and historical contexts.
Our theoretical analysis characterizes target excess risk and
the role of target-specific correction.
Experiments on four datasets demonstrate competitive forecasting
performance, while source-corruption studies highlight the
contributions of group-level and within-group weighting.

Future work will explore three directions.
First, we will examine Transformer-based forecasting backbones
such as PatchTST and extend HiCT to
multi-step prediction.
Second, inspired by online forecasting
methods~\citep{zhang2023onenet}, we will investigate updates
to the weighting parameters as target dynamics evolve.
Finally, we will explore efficient bilevel
optimization~\citep{ji2021bilevel} and predictor refitting
to support larger source libraries.

\bibliography{HiCT}

\appendix
\numberwithin{equation}{section}

\section*{Appendix}

Appendix~\ref{app:A} presents the algorithm and optimization details.
Appendix~\ref{app:theory} provides the theoretical results and proofs.
Appendix~\ref{app:C} describes the datasets and group construction,
and Appendix~\ref{app:weight-visualization} presents heatmaps
of the learned transferability weights.

\section{Algorithm Details}
\label{app:A}

\paragraph{Notation.}
Parenthesized superscripts identify groups, with $(0)$
denoting the target group.
Square-bracketed superscripts identify optimization steps:
$\boldsymbol\Theta^{[m]}$ denotes the weighting parameters
after $m$ updates, whereas $\widetilde{\mathbf W}_i^{[m]}
=\widetilde{\mathbf W}_i(\boldsymbol\Theta^{[m]})$ denotes
the prediction parameters after $m$ updates for validation
query $i$ in the iterative formulation of
Appendix~\ref{app:weight-optimization}.
Thus, $[0]$ indicates initialization.
Let $M\geq1$ be the maximum number of weighting-parameter
updates and $\eta_\Theta>0$ their learning rate.

\paragraph{Validation queries and weight learning.}
We use the validation set
$\mathcal I=\{1,\ldots,T_{\mathrm{val}}\}$ defined in
Section~\ref{section3.4}.
For query $i$, the forecasting origin is
$V_i=T_0-T_{\mathrm{val}}+i-1$.
Let
$\mathbf u_i=\operatorname{vec}(\mathcal C_{V_i}^{(0)})
\in\mathbb R^{Ld}$
and
$\mathbf v_i=\mathbf x_{V_i+1}^{(0)}\in\mathbb R^d$
denote the vectorized context and held-out response,
respectively.
Here, $\operatorname{vec}$ stacks the context entries
in the order used by the predictor.
The conditioning index $i$ in
$\pi_{ks\mid i}=\alpha_{k\mid i}\beta_{s\mid k,i}$
identifies the query.
Weights are normalized over its eligible source library
$\mathcal U_i$, which is fixed across learning iterations.
The corresponding predictor uses preceding target samples
in $\mathcal D_{<i}$ and source contexts in $\mathcal U_i$;
the held-out response is excluded from this fit.

For weight learning, we use a linear predictor.
Its fitted coefficient matrix
$\mathbf W_i(\boldsymbol\Theta^{[m]})
\in\mathbb R^{Ld\times d}$
corresponds to
$\widetilde{\mathbf W}_i(\boldsymbol\Theta)$
in the main text under this parameterization after $m$ updates.
Differentiation through an exact fit requires the selected
solution map to be differentiable with respect to
$\boldsymbol\Theta$.
We denote the predictor's mean squared error across
the $d$ held-out latent coordinates by
$\ell_i(\boldsymbol\Theta)$.
At each iteration, Adam updates the weighting parameters
using the full-validation objective
$\mathcal L_{\mathrm{meta}}$ in
\eqref{eq:meta-transfer-objective}, differentiating through
the query-specific predictor fits.
The validation set remains unchanged throughout learning.
Appendix~\ref{app:weight-optimization} describes the
alternative formulation based on a finite sequence of
differentiable predictor updates.

\paragraph{Rolling prediction.}
Algorithm~\ref{alg:hict} describes the linear-predictor
implementation with rolling refits.
After learning, we fix $\widehat{\boldsymbol\Theta}$
and generate source weights for each new target context.
For a test forecasting origin $t\geq T_0$, let
$\mathbf u_t=\operatorname{vec}(\mathcal C_t^{(0)})$.
The eligible source library is
$\mathcal U_{t-T_0}^{\mathrm{test}}$, and
$\pi_{ks,t}$ denotes the weight assigned to source
context $(k,s)$ at that origin.

The refitted test predictor includes a coefficient matrix
$\widetilde{\mathbf W}_t\in\mathbb R^{Ld\times d}$
and an intercept
$\widetilde{\mathbf b}_t\in\mathbb R^d$.
Unlike the validation predictor used for weight learning,
this predictor therefore includes an intercept.
When target-specific correction is enabled, its coefficient
and intercept corrections are
$\widehat{\boldsymbol\Delta}_t\in\mathbb R^{Ld\times d}$
and $\widehat{\mathbf c}_t\in\mathbb R^d$.
The configured correction multiplier is denoted by
$\gamma_{\mathrm{corr}}$, distinguishing it from
the normalized source-loss coefficient $\gamma$
in the risk analysis.
The latent forecast $\widehat{\mathbf x}_{t+1}^{(0)}$
is reconstructed as
$\widehat{\mathbf y}_{t+1}^{(0)}$
in the target observation space.
The recurrent implementation uses a fixed prediction
head during testing.

\begin{algorithm}[H]
	\caption{Learning Transferable Historical Contexts}
	\label{alg:hict}
	\begin{algorithmic}[1]
		\Require Target and source histories
		$\{\mathbf y_t^{(k)}\}_{t=1}^{T_k}$, $k=0,\ldots,K$;
		context length $L$, stride $h$, and latent dimension $d$;
		maximum learning iterations $M$ and learning rate
		$\eta_\Theta$; validation length $T_{\mathrm{val}}$;
		validation-selected penalty, weighting, and correction settings.
		\Ensure Target forecasts $\widehat{\mathbf y}_{t+1}^{(0)}$.
		
		\State Fit and freeze standardization, PCA, and Procrustes
		alignment using training data;
		\State Construct target samples, the validation set
		$\mathcal I$, and source libraries using the prescribed cutoffs;
		\State Initialize the weighting parameters
		$\boldsymbol\Theta^{[0]}$;
		
		\For{$m=0,\ldots,M-1$}
		\For{$i\in\mathcal I$}
		\State Compute
		$\pi_{ks\mid i}=\alpha_{k\mid i}\beta_{s\mid k,i}$
		using $\boldsymbol\Theta^{[m]}$ over $\mathcal U_i$;
		\State Obtain
		$\mathbf W_i(\boldsymbol\Theta^{[m]})$
		by differentiable weighted predictor fitting,
		using target samples in $\mathcal D_{<i}$
		and source contexts in $\mathcal U_i$;
		\State Compute
		$\ell_i(\boldsymbol\Theta^{[m]})
		\gets d^{-1}
		\|\mathbf v_i-
		\mathbf W_i(\boldsymbol\Theta^{[m]})^\top
		\mathbf u_i\|_2^2$;
		\EndFor
		\State Compute
		$\mathcal L_{\mathrm{meta}}(\boldsymbol\Theta^{[m]})
		\gets T_{\mathrm{val}}^{-1}
		\sum_{i\in\mathcal I}\ell_i(\boldsymbol\Theta^{[m]})$;
		\State Obtain $\boldsymbol\Theta^{[m+1]}$
		by an Adam update with learning rate $\eta_\Theta$,
		differentiating through the predictor fits;
		\EndFor

\State Freeze
$\widehat{\boldsymbol\Theta}\gets\boldsymbol\Theta^{[M]}$
and disable dropout;

\For{each test prediction context $\mathcal C_t^{(0)}$}
\State Compute context-specific weights $\pi_{ks,t}$
using $\widehat{\boldsymbol\Theta}$
over $\mathcal U_{t-T_0}^{\mathrm{test}}$;
\State Apply the selected beta temperature
and retained-mass rule;
\State Normalize the retained nonnegative source
weights to sum to one;
\State Refit the Stage~1 predictor
$(\widetilde{\mathbf W}_t,\widetilde{\mathbf b}_t)$
using the selected penalty;
\State Set
$(\widehat{\boldsymbol\Delta}_t,\widehat{\mathbf c}_t)
\gets(\mathbf0,\mathbf0)$;
\If{target-specific correction is enabled}
\State Fit
$(\widehat{\boldsymbol\Delta}_t,\widehat{\mathbf c}_t)$
to target fitting residuals;
\EndIf
\State Set
$\mathbf u_t\gets\operatorname{vec}(\mathcal C_t^{(0)})$;
\State Compute
$\widehat{\mathbf x}_{t+1}^{(0)}
\gets\widetilde{\mathbf b}_t
+\widetilde{\mathbf W}_t^\top\mathbf u_t
+\gamma_{\mathrm{corr}}
(\widehat{\mathbf c}_t+
\widehat{\boldsymbol\Delta}_t^\top\mathbf u_t)$;
\State Reconstruct $\widehat{\mathbf y}_{t+1}^{(0)}$
through the target PCA decoder and apply
the configured prediction floor;
\EndFor

\State \Return forecasts
$\widehat{\mathbf y}_{t+1}^{(0)}$ for all test origins $t$.
\end{algorithmic}
\end{algorithm}

\subsection{Rolling validation and optimization details}
\label{app:weight-optimization}

The validation length $T_{\mathrm{val}}$ is a positive integer
chosen so that every validation query has a length-$L$ context
and a nonempty set $\mathcal D_{<i}$ of preceding target samples
for predictor fitting.
Every weighting-parameter update uses all $T_{\mathrm{val}}$
validation responses.
All source libraries used for weight normalization are nonempty.
Any retained-mass filtering preserves a positive total weight.
The penalty coefficients and stopping tolerance are nonnegative,
and both learning rates are positive.
The shared prediction-parameter initialization
$\mathbf W_{\mathrm{init}}$ is shared across all predictor fits
and is independent of $\boldsymbol\Theta$.
Any data-dependent initialization uses only observations available
at the earliest rolling-query origin.

\paragraph{Eligible source contexts and normalized weights.}
For each historical query
${\mathcal C_{V_i}^{(0)}}$,
the eligible source library $\mathcal U_i$ is defined by
\eqref{eq:available-source-library}.
Define
\begin{equation*}
	\begin{aligned}
		\mathcal S_{k\mid i}
		&=\bigl\{s\in\{1,\ldots,S_k\}:(k,s)\in\mathcal U_i\bigr\},
		\qquad k=1,\ldots,K,\\
		\mathcal K_i
		&=\bigl\{k\in\{1,\ldots,K\}:|\mathcal S_{k\mid i}|>0\bigr\}.
	\end{aligned}
\end{equation*}
For $k\in\mathcal K_i$, the group summary averages the encoded
contexts indexed by $\mathcal S_{k\mid i}$ only.
The group-level softmax is evaluated over $\mathcal K_i$,
and each within-group softmax is evaluated over $\mathcal S_{k\mid i}$.
Thus, unavailable contexts are excluded before constructing
group summaries and computing either normalization.
Groups outside $\mathcal K_i$ receive zero group weight and zero joint weights, and no within-group normalization is performed for these groups.
The resulting joint weights satisfy
\begin{equation*}
	\begin{aligned}
		\pi_{ks\mid i}(\boldsymbol\Theta)
		&=\pi_{ks}\!\left(
		{\mathcal C_{V_i}^{(0)}};
		\boldsymbol\Theta\right)\geq0,
		&& (k,s)\in\mathcal U_i,\\
		\sum_{(k,s)\in\mathcal U_i}
		\pi_{ks\mid i}(\boldsymbol\Theta)&=1.
	\end{aligned}
\end{equation*}
Weights outside $\mathcal U_i$ are zero.
The dependence of the weighting function on the eligible library is suppressed in the notation.
The held-out response
${\mathbf x_{V_i+1}^{(0)}}$
is excluded from the weighting inputs and the predictor fit.

\paragraph{Prediction-parameter fitting.}
The predictor-fitting target loss averages over the nonempty
set $\mathcal D_{<i}$ of preceding target context--response pairs:
\begin{equation*}
	L_{0,<i}(\mathbf W)
	=\frac{1}{|\mathcal D_{<i}|}
	\sum_{(\mathcal C,\mathbf x)\in\mathcal D_{<i}}
	\ell\!\left(f_{\mathbf W}(\mathcal C),\mathbf x\right),
	\qquad i=1,\ldots,T_{\mathrm{val}}.
\end{equation*}
The regularized predictor-fitting objective is
\begin{equation*}
	\mathcal L_i^{\mathrm{in}}(\mathbf W;\boldsymbol\Theta)
	=L_{0,<i}(\mathbf W)
	+\lambda_{\mathrm{tr}}\sum_{(k,s)\in\mathcal U_i}
	\pi_{ks\mid i}(\boldsymbol\Theta)L_{ks}(\mathbf W)
	+\lambda_{\mathbf W}\mathcal J(\mathbf W).
\end{equation*}
Assuming a minimizer exists, a selected minimizer
$\widetilde{\mathbf W}_i(\boldsymbol\Theta)$ defines the ideal
validation objective in \eqref{eq:meta-transfer-objective}.

For computation, let $\mathcal T_{i,\eta_{\mathrm{in}}}$ denote
the chosen prediction-parameter update with step size
$\eta_{\mathrm{in}}$.
Starting from
$\mathbf W_i^{[0]}(\boldsymbol\Theta)=\mathbf W_{\mathrm{init}}$,
we compute
\begin{equation*}
	\mathbf W_i^{[m+1]}(\boldsymbol\Theta)
	=\mathcal T_{i,\eta_{\mathrm{in}}}\!\left(
	\mathbf W_i^{[m]}(\boldsymbol\Theta);\boldsymbol\Theta\right),
	\qquad m=0,\ldots,M-1.
\end{equation*}
For the differentiation below, the update map is assumed
differentiable with respect to its arguments along the
unrolled trajectory.
When the predictor-fitting objective is twice continuously
differentiable in $(\mathbf W,\boldsymbol\Theta)$,
a gradient descent update is
\begin{equation*}
	\mathcal T_{i,\eta_{\mathrm{in}}}(\mathbf W;\boldsymbol\Theta)
	=\mathbf W-\eta_{\mathrm{in}}
	\nabla_{\mathbf W}\mathcal L_i^{\mathrm{in}}(\mathbf W;\boldsymbol\Theta).
\end{equation*}
For a nonsmooth regularizer, an appropriate update and
differentiation rule must be specified separately;
ordinary gradient descent is not assumed.

\paragraph{Validation-based weighting-parameter updates.}
For the finite-step variant, the $m$-step estimates define
the full-validation objective:
\begin{equation*}
	\mathcal L_{\mathrm{meta}}^{[m]}(\boldsymbol\Theta)
	=\frac{1}{T_{\mathrm{val}}}
	\sum_{i=1}^{T_{\mathrm{val}}}
	\ell\!\left(
	f_{\mathbf W_i^{[m]}(\boldsymbol\Theta)}
	({\mathcal C_{V_i}^{(0)}}),
	{\mathbf x_{V_i+1}^{(0)}}
	\right).
\end{equation*}
Assuming differentiability along the unrolled trajectories,
we compute the gradient of this finite-step objective by backpropagating through all $M$ prediction-parameter updates.
Even when the validation objective based on an exact
predictor fit is differentiable, its gradient need not
coincide with this finite-step gradient.
At learning iteration $m$, an update takes the form
\begin{equation*}
	\boldsymbol\Theta^{[m+1]}
	=\boldsymbol\Theta^{[m]}
	-\eta_\Theta\nabla_{\boldsymbol\Theta}
	\mathcal L_{\mathrm{meta}}^{[m]}
	\!\left(\boldsymbol\Theta^{[m]}\right).
\end{equation*}
In the finite-step variant, $\boldsymbol\Theta$ stores
the current weighting-parameter iterate.
The code-based algorithm instead differentiates through
predictor fits and uses its fixed Adam update budget.

For the finite-step variant, we perform at most
$M$ weighting-parameter updates.
At each iteration, the loss
$v_m=\mathcal L_{\mathrm{meta}}^{[m]}
(\boldsymbol\Theta^{[m]})$ is evaluated before updating
the parameters.
For $m\geq1$, optimization stops before the next update if
$|v_m-v_{m-1}|\leq\epsilon_{\mathrm{tol}}$.
The returned $\widehat{\boldsymbol\Theta}$ is the current
iterate when this condition is met or the update budget is exhausted.

\paragraph{Current-origin transfer and prediction.}
At the current forecasting origin $\tau_{T_0}^{(0)}$, define
\begin{equation*}
	\mathcal U_0
	=\left\{(k,s):
	1\leq k\leq K,\;1\leq s\leq S_k,\;
	\tau_{t_{ks}+1}^{(k)}\leq\tau_{T_0}^{(0)}
	\right\}.
\end{equation*}
Applying the same eligible-context restrictions to group
summaries and both normalizations gives
\begin{equation}
	\widehat\pi_{ks}
	=\pi_{ks}\!\left(\mathcal C_0;\widehat{\boldsymbol\Theta}\right),
	\qquad (k,s)\in\mathcal U_0,
	\label{eq:final-transferability}
\end{equation}
where $\widehat\pi_{ks}\geq0$ and
$\sum_{(k,s)\in\mathcal U_0}\widehat\pi_{ks}=1$.
Unavailable contexts receive zero weight.
Using these weights and all target forecasting samples
available at the current origin, we fit $\widetilde{\mathbf W}$
using \eqref{eq:weighted-transfer}.
If target-specific correction is enabled, we estimate
$\widehat{\boldsymbol\Delta}$ using \eqref{eq:target-debiasing};
otherwise, we set $\widehat{\boldsymbol\Delta}=\mathbf0$.
The final estimator is
$\widehat{\mathbf W}=\widetilde{\mathbf W}+\widehat{\boldsymbol\Delta}$,
and its forecast is reconstructed using
\eqref{eq:forecast-reconstruction}.
This is the unscaled correction defined in
Section~\ref{section3.3}. Algorithm~\ref{alg:hict} also allows
an implementation-specific correction multiplier and prediction
floor; these operations are separate from the estimator analyzed
in Proposition~\ref{prop:2}.

\section{Theoretical Properties}
\label{app:theory}

Proposition~\ref{prop:1} bounds the target excess risk
of the regularized transfer estimator under a uniform
deviation condition.
Proposition~\ref{prop:2} establishes an empirical guarantee
for target-specific adaptation.
We illustrate weight concentration with a weighted-mean model and the effect of the correction penalty with a simplified ridge model. These calculations explain statistical trade-offs under explicit simplifying assumptions.

\subsection{Context-weighted transfer estimation}
\label{app:transfer-theory}

Let $\mathcal E$ be a finite-dimensional real Euclidean space.
For the transfer-risk analysis, consider a fixed nonempty
compact parameter set $\mathcal W\subseteq\mathcal E$.
The representation maps are treated as deterministic,
and the index set of available source contexts is fixed
and nonempty.
This analysis does not account for uncertainty from
estimating the representation maps.

We use $L$ to denote an empirical loss, $R$ to denote
a population risk, and $\mathcal J$ to denote
a regularization penalty.
The Stage~1 penalty $\mathcal J(\mathbf W)$ controls
the complexity of the forecasting model.
We write $Q_{\boldsymbol\pi}$ for the original regularized
empirical objective and $\overline Q_{\boldsymbol\pi}$
for its normalized version.

Assume that the empirical losses are measurable functions
of the data and are finite and continuous on $\mathcal W$
almost surely.
The corresponding population risks are assumed to be
finite and continuous on $\mathcal W$.
The regularizer $\mathcal J$ is deterministic, finite,
nonnegative, and continuous on $\mathcal W$.
We restrict our analysis to transferred estimators
$\widetilde{\mathbf W}\in\mathcal W$.
All estimators and generalization-gap events considered below are assumed measurable.

Let $L_0(\mathbf W)$ denote the empirical target loss and
$L_{ks}(\mathbf W)$ the prediction loss for the source
context--response pair indexed by $(k,s)$.
For each fixed $\mathbf W\in\mathcal W$, define
the corresponding population risks as
\begin{equation*}
	R_0(\mathbf W)=\mathbb E[L_0(\mathbf W)],
	\qquad
	R_{ks}(\mathbf W)=\mathbb E[L_{ks}(\mathbf W)].
\end{equation*}
The expectations are taken over the time-series data
under the fixed setup described above.
Here, $R_0$ is the population counterpart of the empirical
loss over historical target forecasting samples.
Relating this risk to prediction at the current forecasting
origin requires additional assumptions on temporal
distribution changes.

We measure the source--target discrepancy by $\delta_{ks}
=\sup_{\mathbf W\in\mathcal W}
\left|R_{ks}(\mathbf W)-R_0(\mathbf W)
\right|$.
Each $\delta_{ks}$ is finite under the compactness and
continuity assumptions above.
A smaller $\delta_{ks}$ indicates closer agreement between
the source-context risk and the target risk over $\mathcal W$.

Let the transferability weights $\boldsymbol\pi=(\pi_{ks})$
satisfy
\begin{equation*}
	\pi_{ks}\geq 0,
	\qquad
	\sum_{k=1}^{K}\sum_{s=1}^{S_k}\pi_{ks}=1,
\end{equation*}
with zero weights assigned to unavailable source contexts.
These are the normalization conditions in
\eqref{normalized_weights}.
For fixed hyperparameters
$\lambda_{\mathrm{tr}},\lambda_{\mathbf W}\geq 0$,
define the context-weighted population risk as
\begin{equation*}
	R_{\boldsymbol\pi}(\mathbf W)
	=
	\frac{
		R_0(\mathbf W)
		+
		\lambda_{\mathrm{tr}}
		\sum_{k=1}^{K}\sum_{s=1}^{S_k}
		\pi_{ks}R_{ks}(\mathbf W)
	}{
		1+\lambda_{\mathrm{tr}}
	}.
\end{equation*}
Its empirical counterpart is
\begin{equation*}
	L_{\boldsymbol\pi}(\mathbf W)
	=
	\frac{
		L_0(\mathbf W)
		+
		\lambda_{\mathrm{tr}}
		\sum_{k=1}^{K}\sum_{s=1}^{S_k}
		\pi_{ks}L_{ks}(\mathbf W)
	}{
		1+\lambda_{\mathrm{tr}}
	}.
\end{equation*}

Let $\mathbf W_0^*$ denote a target population risk minimizer,
\begin{equation*}
	\mathbf W_0^*
	\in
	\arg\min_{\mathbf W\in\mathcal W}
	R_0(\mathbf W).
\end{equation*}
Such a minimizer exists because $\mathcal W$ is nonempty
and compact and $R_0$ is continuous on $\mathcal W$.

For fixed transferability weights, the regularized empirical
objective in Stage~1 is
\begin{equation*}
	Q_{\boldsymbol\pi}(\mathbf W)
	=
	L_0(\mathbf W)
	+
	\lambda_{\mathrm{tr}}
	\sum_{k=1}^{K}\sum_{s=1}^{S_k}
	\pi_{ks}L_{ks}(\mathbf W)
	+
	\lambda_{\mathbf W}\mathcal J(\mathbf W).
\end{equation*}
Dividing this objective by $1+\lambda_{\mathrm{tr}}$
gives the normalized objective,
with normalized regularization coefficient $\rho:=\lambda_{\mathbf W}/(1+\lambda_{\mathrm{tr}})$,
\begin{equation*}
	\begin{aligned}
		\overline Q_{\boldsymbol\pi}(\mathbf W)
		&=
		\frac{Q_{\boldsymbol\pi}(\mathbf W)}{1+\lambda_{\mathrm{tr}}}
		=
		L_{\boldsymbol\pi}(\mathbf W)
		+
		\rho\mathcal J(\mathbf W).
	\end{aligned}
\end{equation*}
This normalization rescales both the loss and the
regularization term.
Since $1+\lambda_{\mathrm{tr}}>0$ is independent of
$\mathbf W$, the original and normalized objectives
have the same minimizers over $\mathcal W$.

To account for approximate optimization, assume that
the transferred estimator $\widetilde{\mathbf W}\in\mathcal W$
satisfies
\begin{equation}
	\overline Q_{\boldsymbol\pi}(\widetilde{\mathbf W})
	\leq
	\inf_{\mathbf W\in\mathcal W}
	\overline Q_{\boldsymbol\pi}(\mathbf W)
	+
	\varepsilon_{\mathrm{opt}},
	\label{eq:approximate-transfer-estimation}
\end{equation}
where $\varepsilon_{\mathrm{opt}}\geq 0$ is finite and
measures the optimization error in the normalized objective.
Exact minimization corresponds to
$\varepsilon_{\mathrm{opt}}=0$.

\paragraph{Proof of Proposition~\ref{prop:1}.}

\begin{proof}
	Let $D_{\boldsymbol\pi}
	=\sum_{k=1}^{K}\sum_{s=1}^{S_k}
	\pi_{ks}\delta_{ks}$, $\gamma
	=\frac{\lambda_{\mathrm{tr}}}{1+\lambda_{\mathrm{tr}}}$
	and recall that
	$\rho=\lambda_{\mathbf W}/(1+\lambda_{\mathrm{tr}})$.
	For any $\mathbf W\in\mathcal W$, the definitions of
	$R_{\boldsymbol\pi}$ and $\delta_{ks}$, together with
	the nonnegativity and normalization of the weights, imply
	\begin{align}
		\left|
		R_{\boldsymbol\pi}(\mathbf W)-R_0(\mathbf W)
		\right|
		=
		\gamma
		\left|
		\sum_{k=1}^{K}\sum_{s=1}^{S_k}
		\pi_{ks}
		\left[
		R_{ks}(\mathbf W)-R_0(\mathbf W)
		\right]
		\right|
		\nonumber &\leq
		\gamma
		\sum_{k=1}^{K}\sum_{s=1}^{S_k}
		\pi_{ks}
		\left|
		R_{ks}(\mathbf W)-R_0(\mathbf W)
		\right|
		\nonumber\\
		&\leq
		\gamma D_{\boldsymbol\pi}.
		\label{eq:proof-population-discrepancy}
	\end{align}
	
	Next, since $\mathbf W_0^*\in\mathcal W$, the approximate
	optimality condition
	\eqref{eq:approximate-transfer-estimation} gives
	\begin{equation*}
		\overline Q_{\boldsymbol\pi}(\widetilde{\mathbf W})
		\leq
		\inf_{\mathbf W\in\mathcal W}
		\overline Q_{\boldsymbol\pi}(\mathbf W)
		+
		\varepsilon_{\mathrm{opt}}
		\leq
		\overline Q_{\boldsymbol\pi}(\mathbf W_0^*)
		+
		\varepsilon_{\mathrm{opt}}.
	\end{equation*}
	Substituting
	$\overline Q_{\boldsymbol\pi}
	=L_{\boldsymbol\pi}+\rho\mathcal J$
	and rearranging yields
	\begin{equation}
		L_{\boldsymbol\pi}(\widetilde{\mathbf W})
		-
		L_{\boldsymbol\pi}(\mathbf W_0^*)
		\leq
		\rho
		\left[
		\mathcal J(\mathbf W_0^*)
		-
		\mathcal J(\widetilde{\mathbf W})
		\right]
		+
		\varepsilon_{\mathrm{opt}}
		\nonumber\leq
		\rho\mathcal J(\mathbf W_0^*)
		+
		\varepsilon_{\mathrm{opt}},
		\label{eq:proof-regularized-optimality}
	\end{equation}
	where the last inequality uses $\rho\geq 0$ and
	$\mathcal J(\widetilde{\mathbf W})\geq 0$.
	
	Applying \eqref{eq:proof-population-discrepancy} at
	$\widetilde{\mathbf W}$ and $\mathbf W_0^*$ gives
	\begin{equation*}
		R_0(\widetilde{\mathbf W})-R_0(\mathbf W_0^*)
		\leq
		R_{\boldsymbol\pi}(\widetilde{\mathbf W})
		-
		R_{\boldsymbol\pi}(\mathbf W_0^*)
		+
		2\gamma D_{\boldsymbol\pi}.
	\end{equation*}
	The generalization-gap bound \eqref{eq:weighted-risk-deviation} also implies
	\begin{equation*}
		R_{\boldsymbol\pi}(\widetilde{\mathbf W})
		-
		R_{\boldsymbol\pi}(\mathbf W_0^*)
		\leq
		L_{\boldsymbol\pi}(\widetilde{\mathbf W})
		-
		L_{\boldsymbol\pi}(\mathbf W_0^*)
		+
		2\varepsilon_{\mathrm{gen}}(\boldsymbol\pi).
	\end{equation*}
	Combining these inequalities with
	\eqref{eq:proof-regularized-optimality}, we obtain
	\begin{equation*}
		R_0(\widetilde{\mathbf W})-R_0(\mathbf W_0^*)
		\leq
		2\gamma D_{\boldsymbol\pi}
		+
		\rho\mathcal J(\mathbf W_0^*)
		+
		2\varepsilon_{\mathrm{gen}}(\boldsymbol\pi)
		+
		\varepsilon_{\mathrm{opt}}.
	\end{equation*}
	Substituting the definitions of $D_{\boldsymbol\pi}$,
	$\gamma$, and $\rho$ proves
	\eqref{prop:context-transfer}.
\end{proof}

Proposition~\ref{prop:1} bounds the target excess risk
by four contributions: weighted source--target discrepancy,
a uniform generalization-gap bound, regularization,
and optimization error.
Shifting weight toward contexts with smaller $\delta_{ks}$
can reduce the discrepancy term.
However, the generalization term may also vary with
the weight allocation.

\paragraph{Data-dependent transferability weights.}
For data-dependent weights, a sufficient condition is that the generalization-gap bound holds simultaneously for every weight vector in a fixed class $\Pi$. This class must contain all admissible weights that the learning procedure may return.
Let
$\varepsilon_{\mathrm{gen}}:\Pi\to[0,\infty)$
be a deterministic finite-valued function.
Under the same fixed setup, suppose that, for some
$\zeta\in(0,1)$,
\begin{equation*}
	\mathbb P\left(
	\sup_{\mathbf W\in\mathcal W}
	\left|
	L_{\boldsymbol\pi}(\mathbf W)
	-
	R_{\boldsymbol\pi}(\mathbf W)
	\right|
	\leq
	\varepsilon_{\mathrm{gen}}(\boldsymbol\pi)
	\quad
	\text{for all }\boldsymbol\pi\in\Pi
	\right)
	\geq 1-\zeta.
\end{equation*}
On this event, the proof of Proposition~\ref{prop:1}
applies to the learned weights
$\widehat{\boldsymbol\pi}\in\Pi$ and any corresponding
estimator satisfying
\eqref{eq:approximate-transfer-estimation}.
Thus, the same risk bound holds with
$\boldsymbol\pi=\widehat{\boldsymbol\pi}$.
For each fixed $\mathbf W\in\mathcal W$,
$R_{\widehat{\boldsymbol\pi}}(\mathbf W)$ denotes
$R_{\boldsymbol\pi}(\mathbf W)$ evaluated at the learned weights.
It need not equal
$\mathbb E[L_{\widehat{\boldsymbol\pi}}(\mathbf W)]$
when the weights depend on the same data.

The simultaneous bound allows the weights to depend on the data; the argument does not establish that this condition holds for the full HiCT procedure.

It retains the assumptions of fixed hyperparameters,
fixed representation maps, and a fixed source-context
index set.
Deriving explicit rates for
$\varepsilon_{\mathrm{gen}}(\boldsymbol\pi)$ requires
additional assumptions on the loss class, the weight
class $\Pi$, and the temporal dependence of the data.

\paragraph{Comparison with uniform weighting.}

A comparison with uniform weighting requires separate control of the weighting-parameter learning problem. Consider a common rolling-validation protocol with the same representation maps, source availability restrictions, and predictor-fitting procedure. If the empirical rolling-validation loss uniformly approximates its finite population expectation over a fixed class containing all possible learned rules and the uniform rule, and the learned rule's empirical loss exceeds that of uniform weighting by at most a comparison tolerance, then its population rolling-prediction risk exceeds that of uniform weighting by at most twice the uniform generalization-gap bound plus that tolerance.

This conditional comparison bounds possible risk deterioration;
it does not guarantee an improvement over uniform weighting.
It concerns the population counterpart of the rolling-validation
objective and does not directly extend to the final refitted
or target-adapted predictor.

\paragraph{Weight concentration and temporal dependence.}
To isolate the effect of weighting, consider scalar source summaries $Z_{ks}=\theta_0+b_{ks}+\epsilon_{ks}$, where $\theta_0$ is the target mean, $b_{ks}$ is a fixed source shift, and the stacked noise vector $\boldsymbol\epsilon$ has zero mean and finite covariance $\Sigma_\epsilon$. Hold the representations and normalized weights $\pi_{ks}=\alpha_k\beta_{s\mid k}$ fixed, and define $\widehat\theta_{\boldsymbol\pi}=\sum_{k,s}\pi_{ks}Z_{ks}$. Then
\begin{equation*}
	\mathbb E\!\left[(\widehat\theta_{\boldsymbol\pi}-\theta_0)^2\right]
	=\left(\sum_{k,s}\pi_{ks}b_{ks}\right)^2
	+\boldsymbol\pi^\top\Sigma_\epsilon\boldsymbol\pi.
\end{equation*}
This identity follows by separating the mean shift from the zero-mean noise.
For independent errors with common variance $\sigma^2>0$,
the variance term reduces to
\begin{equation*}
	\boldsymbol\pi^\top\Sigma_\epsilon\boldsymbol\pi
	=\sigma^2\sum_k\alpha_k^2\sum_s\beta_{s\mid k}^2.
\end{equation*}

In particular, assigning all weight to a single source context with a
small  absolute mean  shift can keep the
squared  bias small, but it makes
$\sum_{k,s}\pi_{ks}^{2}=1$  and yields variance    $\sigma^{2}$,
the largest possible variance among normalized nonnegative weights
in this independent, homoscedastic model.

Under the same noise assumptions, assigning positive weight
to more than one context strictly reduces the variance
relative to assigning all weight to a single context.
Thus, concentrating on the source context with the smallest
absolute mean shift need not minimize mean squared error.

With temporally dependent or overlapping contexts,
cross-context error  covariances also matter,
so  weight  concentration alone does not    determine
the variance.
This is an illustration for a weighted-mean estimator
with fixed weights, not a risk bound for the
full HiCT predictor.
In particular, this calculation does not provide the generalization-gap bound required by Proposition~\ref{prop:1}.
HiCT learns its weighting function from target validation loss,
without explicitly minimizing  this illustrative variance.

\subsection{Target-specific adaptation}
\label{app:adaptation-theory}

We next analyze the target-specific correction while
holding the transferred estimator
$\widetilde{\mathbf W}$ fixed.
For this stage, $L_0$ is defined on the ambient parameter
space $\mathcal E$, and the correction is optimized
over $\mathcal E$.
Let $\mathcal J_\Delta$ denote the regularization penalty
on the correction.
For $\lambda_\Delta\geq 0$, define the second-stage
objective as
\begin{equation*}
	Q_{\mathrm{ad}}(\boldsymbol\Delta)
	=
	L_0\!\left(
	\widetilde{\mathbf W}+\boldsymbol\Delta
	\right)
	+
	\lambda_\Delta
	\mathcal J_\Delta(\boldsymbol\Delta).
\end{equation*}
The correction is obtained from
\begin{equation}
	\widehat{\boldsymbol\Delta}
	\in
	\arg\min_{\boldsymbol\Delta\in\mathcal E}
	Q_{\mathrm{ad}}(\boldsymbol\Delta),
	\label{eq:appendix-target-correction}
\end{equation}
and the final estimator is
\begin{equation*}
	\widehat{\mathbf W}
	=
	\widetilde{\mathbf W}
	+
	\widehat{\boldsymbol\Delta}.
\end{equation*}

\begin{proposition}[Stage 2: Empirical guarantee for target-specific adaptation]
	\label{prop:2}
	Suppose that
	$L_0(\widetilde{\mathbf W}+\boldsymbol\Delta)$
	and $\mathcal J_\Delta(\boldsymbol\Delta)$
	are finite, convex, and continuous functions of
	$\boldsymbol\Delta$ on $\mathcal E$.
	Assume that
	$\mathcal J_\Delta(\boldsymbol\Delta)\geq 0$
	for all $\boldsymbol\Delta\in\mathcal E$,
	$\mathcal J_\Delta(\mathbf 0)=0$,
	and $\lambda_\Delta\geq 0$.
	Suppose that
	\eqref{eq:appendix-target-correction}
	admits a minimizer
	$\widehat{\boldsymbol\Delta}$.
	Then the following properties hold.
	
	\begin{enumerate}
		
		\item[(i)]
		{The regularized empirical target objective and the unpenalized loss do not increase relative to the zero correction:}
		\begin{equation}
			{\displaystyle L_0(\widehat{\mathbf W})
				\leq L_0(\widehat{\mathbf W})
				+\lambda_\Delta\mathcal J_\Delta(\widehat{\boldsymbol\Delta})
				\leq L_0(\widetilde{\mathbf W}).}
			\label{eq:regularized-adaptation-improvement}
		\end{equation}

		\item[(ii)]
		The zero correction is optimal if and only if
		\begin{equation}
			\mathbf 0
			\in
			\partial L_0(\widetilde{\mathbf W})
			+
			\lambda_\Delta
			\partial\mathcal J_\Delta(\mathbf 0),
			\label{eq:zero-correction-condition}
		\end{equation}
		where $\partial$ denotes the convex subdifferential
		on $\mathcal E$.
		If $Q_{\mathrm{ad}}$ is strictly convex,
		its minimizer is unique.
		In this case,
		\eqref{eq:zero-correction-condition}
		is equivalent to
		$\widehat{\boldsymbol\Delta}=\mathbf 0$.
	\end{enumerate}
\end{proposition}

The guarantee in part~(i) concerns the empirical target
loss used to estimate the correction.
It applies to an exact minimizer of the second-stage
objective and does not imply a reduction in population
or test risk.

\paragraph{Proof of Proposition~\ref{prop:2}.}

\begin{proof}
	\textnormal{(i)}
	The zero correction is feasible for
	\eqref{eq:appendix-target-correction}.
	By the optimality of
	$\widehat{\boldsymbol\Delta}$ and the condition
	$\mathcal J_\Delta(\mathbf 0)=0$,
	\begin{equation*}
		Q_{\mathrm{ad}}(\widehat{\boldsymbol\Delta})
		\leq
		Q_{\mathrm{ad}}(\mathbf 0)
		=
		L_0(\widetilde{\mathbf W}).
	\end{equation*}
	Substituting the definition of $Q_{\mathrm{ad}}$
	and using
	$\widehat{\mathbf W}
	=\widetilde{\mathbf W}+\widehat{\boldsymbol\Delta}$
	proves
	\eqref{eq:regularized-adaptation-improvement}.
	Since
	$\lambda_\Delta\mathcal J_\Delta
	(\widehat{\boldsymbol\Delta})\geq 0$,
	it also follows that
	\begin{equation*}
		L_0(\widehat{\mathbf W})
		\leq
		L_0(\widetilde{\mathbf W}).
	\end{equation*}
	
	\textnormal{(ii)}
	The objective $Q_{\mathrm{ad}}$ is convex.
	The convex optimality condition therefore gives
	\begin{equation*}
		\mathbf 0
		\in
		\arg\min_{\boldsymbol\Delta\in\mathcal E}
		Q_{\mathrm{ad}}(\boldsymbol\Delta)
		\quad\Longleftrightarrow\quad
		\mathbf 0\in\partial Q_{\mathrm{ad}}(\mathbf 0).
	\end{equation*}
	The subdifferential sum rule applies because both
	terms in the objective are finite, convex, and continuous
	on $\mathcal E$.
	Together with translation by the fixed parameter
	$\widetilde{\mathbf W}$, it yields
	\begin{align*}
		\partial Q_{\mathrm{ad}}(\mathbf 0)
		&=
		\left.
		\partial_{\boldsymbol\Delta}
		\left[
		L_0(\widetilde{\mathbf W}+\boldsymbol\Delta)
		\right]
		\right|_{\boldsymbol\Delta=\mathbf 0}
		+
		\lambda_\Delta
		\partial\mathcal J_\Delta(\mathbf 0)
		\\
		&=
		\partial L_0(\widetilde{\mathbf W})
		+
		\lambda_\Delta
		\partial\mathcal J_\Delta(\mathbf 0).
	\end{align*}
	Thus, the zero correction is optimal if and only if
	\eqref{eq:zero-correction-condition} holds.
	
	Finally, strict convexity implies that the minimizer,
	whose existence is assumed, is unique.
	Under this additional condition, zero is optimal
	if and only if
	$\widehat{\boldsymbol\Delta}=\mathbf 0$.
\end{proof}

Proposition~\ref{prop:2} concerns the empirical target
loss used to estimate the correction.
The stated guarantee applies to an exact minimizer
of the second-stage objective.
It does not imply a reduction in population or
future prediction risk.
Under strict convexity,
\eqref{eq:zero-correction-condition} characterizes
when the transferred estimator remains unchanged.

\paragraph{Bias--variance trade-off in penalized target correction.}
To explain why the empirical guarantee in Proposition~\ref{prop:2} need not imply lower prediction risk, consider one output coordinate with $n$ target samples and $p$ predictor coefficients, where $n\geq p\geq1$. Let the fixed design $X\in\mathbb R^{n\times p}$ satisfy $X^\top X/n=I_p$, where $I_p$ is the identity matrix, and let $\mathbf y=X\mathbf w_0+\boldsymbol\epsilon$ where $\mathbf w_0\in\mathbb R^p$ is the target coefficient vector and $\boldsymbol\epsilon\in\mathbb R^n$ has zero mean and finite covariance. Treat the transferred coefficient vector $\widetilde{\mathbf w}$ as fixed and define its remaining error by $\mathbf d=\mathbf w_0-\widetilde{\mathbf w}$. Write $\boldsymbol\xi=X^\top\boldsymbol\epsilon/n$ and $V=\mathbb E\|\boldsymbol\xi\|_2^2$.

For a fixed penalty $\lambda=\lambda_\Delta>0$, the penalized
correction and its expected squared parameter error satisfy
\begin{equation*}
	\begin{aligned}
		\widehat{\boldsymbol\delta}_\lambda
		&=\arg\min_{\boldsymbol\delta\in\mathbb R^p}
		\left\{\frac{\|\mathbf y-X(\widetilde{\mathbf w}+\boldsymbol\delta)\|_2^2}{2n}
		+\frac{\lambda}{2}\|\boldsymbol\delta\|_2^2\right\}
		=\frac{\mathbf d+\boldsymbol\xi}{1+\lambda},\\
		\mathbb E\|\widetilde{\mathbf w}+\widehat{\boldsymbol\delta}_\lambda-\mathbf w_0\|_2^2
		&=\frac{\lambda^2\|\mathbf d\|_2^2+V}{(1+\lambda)^2}.
	\end{aligned}
\end{equation*}

The closed-form correction follows from the ridge normal equations.
Substituting it into the estimation error gives the risk identity
because $\mathbb E\boldsymbol\xi=\mathbf0$.
For a test covariate independent of the training noise,
with second-moment matrix $I_p$, assume that the conditional
mean of the test response follows the same linear model.
Then the expected squared parameter error equals the expected
squared prediction error relative to that mean.

Omitting correction leaves error $\|\mathbf d\|_2^2$. Hence the correction strictly improves this risk precisely when $V<(1+2\lambda)\|\mathbf d\|_2^2$.
Increasing the penalty suppresses correction noise but also
leaves more of the transfer error uncorrected.
If $\mathbf d=\mathbf0$, correction cannot strictly reduce
this risk and increases it whenever $V>0$. The shrinkage here is induced by the ridge penalty, without an additional blend parameter. This calculation assumes a fixed transferred estimator and an identity Gram matrix. If the transferred estimator reuses the correction responses, a cross term can remain, and the displayed identity requires additional justification.
\section{Experimental Details}
\label{app:C}

This section describes the datasets, group construction, and preprocessing used in the experiments in Section~\ref{section:4}.

\subsection{Datasets and Group Construction}
\label{app:datasets}

The lengths reported in the dataset tables include training, validation, and test observations; in Section~\ref{section3}, $T_0$ denotes the target history available before the first test forecast.

Each benchmark contains a short-history target group and multiple source
groups with longer histories. Group construction precedes forecasting,
and source contexts are filtered by calendar-time availability at each
target forecasting origin.

\paragraph{Preprocessing and reconstruction.}
\label{app:preprocessing}
Each variable is transformed as specified below and standardized using statistics fitted through the target training cutoff. The source transformations use only source observations available by that cutoff. These fitted transformations remain fixed during validation and testing. Source contexts are filtered by response availability at each forecasting origin.

For HiCT, the representation and source-to-target maps in Section~\ref{section3.2} are fitted on training observations. The target decoder maps latent forecasts back through the target representation. For the PCA representation, this step applies the target loading matrix and restores the fitted centering vector. We evaluate MSE and MAE against the full standardized target observations, rather than their low-dimensional projections. Thus the decoder used for these metrics stops in standardized observation space; recovering original units would additionally invert standardization and the dataset-specific transformation. This representation step is specific to the relevant HiCT configurations and does not imply that every baseline uses PCA.

\subsubsection{Stock}
\label{app:C1}

The Stock dataset contains daily realized-volatility series from five equity sectors, obtained through Wharton Research Data Services (\url{https://wrds-www.wharton.upenn.edu}).
Consumer Staples serves as the target group, with 56 stocks observed over 277 trading days.
Financials, Health Care, Industrials, and Information Technology serve as the four source groups, with 134, 103, 112, and 126 stocks, respectively.
Table~\ref{tab:data-summary} summarizes the group dimensions and observation periods.

\begin{table}[H]
	\centering
	\normalsize
	\setlength{\tabcolsep}{4pt}
	\renewcommand{\arraystretch}{1.1}
	\caption{Realized-volatility group construction. $N_k$ is the number of
		stocks and $T_k$ is the number of common daily observations.}
	\label{tab:data-summary}
	\begin{tabular}{llrrll}
		\toprule
		Sector & Role & $N_k$ & $T_k$ & Start date & End date \\
		\midrule
		Consumer Staples & Target & {56} & 277 & {2024-04-05} & 2025-08-15 \\
		Financials & Source & {134} & 1,100 & 2021-01-25 & 2025-08-15 \\
		Health Care & Source & {103} & 1,143 & 2020-09-28 & 2025-08-15 \\
		Industrials & Source & {112} & {815} & 2022-04-06 & 2025-08-15 \\
		Information Technology & Source & {126} & 760 & 2022-04-25 & 2025-08-15 \\
		\bottomrule
	\end{tabular}
\end{table}

We retain stocks with at least 60 intraday observations on each included trading day.
For stock $i$ on day $t$, the $j$th five-minute log return is
\begin{equation*}
	r_{i,t,j}
	=
	\log P_{i,t,j}-\log P_{i,t,j-1},
\end{equation*}
where $P_{i,t,j}$ denotes the price at the $j$th five-minute observation.
Returns are constructed within the regular trading session from 9:30 to 16:00, excluding overnight returns.
The daily realized variance is
\begin{equation*}
	RV_{i,t}
	=
	\sum_{j=1}^{M_{i,t}}r_{i,t,j}^{2},
\end{equation*}
where $M_{i,t}$ is the number of available intraday returns.
We split the target history chronologically into 166 training, 55 validation, and 56 test observations. We log-transform $RV_{i,t}$ and standardize each series using its training-period mean and standard deviation.

\subsubsection{NY Taxi}
\label{app:C2}

We use hourly pickup counts derived from the NYC Taxi and Limousine Commission trip records (\url{https://www.nyc.gov/site/tlc/about/tlc-trip-record-data.page}) to evaluate demand forecasting with a short target history. The target and source series are aligned by their observation timestamps, with source availability enforced at each forecasting origin.

The panel records hourly pickup counts by taxi zone. Staten Island is the target group, and the other four boroughs provide 242 source zones (Table~\ref{tab:ny-taxi-details}). The target history is split chronologically into 864 training, 288 validation, and 288 test observations. We apply a $\log(1+x)$ transformation to accommodate zero counts, then standardize each zone series using its training-period mean and standard deviation.

\begin{table}[H]
	\centering
	\normalsize
	\setlength{\tabcolsep}{4pt}
	\renewcommand{\arraystretch}{1.1}
	\caption{NY Taxi group construction. $N_k$ is the number of zones and $T_k$ is the number of hourly observations.}
	\label{tab:ny-taxi-details}
	\begin{tabular}{llrrll}
		\toprule
		Borough & Role & $N_k$ & $T_k$ & Start date & End date \\
		\midrule
		Staten Island & Target & 20 & 1,440 & 2025-11-02 & 2025-12-31 \\
		Bronx & Source & 43 & 8,760 & 2025-01-01 & 2025-12-31 \\
		Brooklyn & Source & 61 & 8,760 & 2025-01-01 & 2025-12-31 \\
		Queens & Source & 69 & 8,760 & 2025-01-01 & 2025-12-31 \\
		Manhattan & Source & 69 & 8,760 & 2025-01-01 & 2025-12-31 \\
		\bottomrule
	\end{tabular}
\end{table}

\subsubsection{PEMS-BAY}
\label{app:C3}

The PEMS-BAY panel contains traffic-speed measurements from 325 sensors sampled every five minutes during June 2017. The source dataset is available at \url{https://zenodo.org/records/5724362}.
Each source group retains 8,640 observations, while the target group retains the final 2,016 observations.
We project the sensor coordinates onto a local Cartesian coordinate system and apply capacity-constrained $K$-means to obtain five geographical groups of 65 sensors each.
Clustering uses fixed initialization settings and does not depend on forecasting outcomes.

The group whose centroid is closest to the centroid of the complete sensor network is designated as the target group. The remaining groups are ordered by centroid longitude and used as source groups.
The target history covers June 24--30, 2017, corresponding to seven days, while each source history covers June 1--30.
We split the target history chronologically into 1,209 training, 403 validation, and 404 test observations. We apply the transformation $x\mapsto\log(1+x)$ to each sensor series, followed by standardization using training-period statistics.

\begin{table}[H]
	\centering
	\normalsize
	\setlength{\tabcolsep}{4pt}
	\renewcommand{\arraystretch}{1.1}
	\caption{PEMS-BAY group construction. Each geographical group contains
		65 sensors.}
	\label{tab:pems-bay}
	\begin{tabular}{llrrl}
		\toprule
		Group & Role & $N_k$ & $T_k$ & Observation period \\
		\midrule
		Region 5 & Target & 65 & {2,016} & {June 24--30, 2017} \\
		Regions 1--4, each & Source & 65 & {8,640} & {June 1--30, 2017} \\
		\bottomrule
	\end{tabular}
\end{table}

\subsubsection{Electricity}
\label{app:C4}

The Electricity panel contains hourly consumption series from 321 customers over 2012--2014, with 26,304 observations per customer. The source dataset is available at \url{https://zenodo.org/records/4656140}.
The target group consists of the final 60 customers, T262--T321.
We rank the remaining 261 customers by the distance of their standardized training-period trajectories from the mean target trajectory and divide this ranking into three groups of 87 customers.

The target retains the final 1,440 hourly observations, which are split chronologically into 864 training, 288 validation, and 288 test observations.
No logarithmic transformation is applied, and each customer series is standardized using training-period statistics.

Source histories are aligned with the target by calendar time.
At each forecasting origin, transfer is restricted to source contexts whose one-step-ahead responses are available.

\begin{table}[H]
	\centering
	\normalsize
	\setlength{\tabcolsep}{4pt}
	\renewcommand{\arraystretch}{1.1}
	\caption{Electricity group construction. $N_k$ is the number of customers and $T_k$ is the number of hourly observations. Source groups are formed by training-period trajectory similarity; the target row retains the original customer identifiers.}
	\label{tab:electricity}
	\begin{tabular}{llrrll}
		\toprule
		Customers & Role & $N_k$ & $T_k$ & Start date & End date \\
		\midrule
		T262--T321 & Target & 60 & 1,440 & 2014-11-02 & 2014-12-31 \\
		Group 1    & Source & 87 & 26,304 & 2012-01-01 & 2014-12-31 \\
		Group 2 & Source & 87 & 26,304 & 2012-01-01 & 2014-12-31 \\
		Group 3 & Source & 87 & 26,304 & 2012-01-01 & 2014-12-31 \\
		\bottomrule
	\end{tabular}
\end{table}

\subsection{Evaluation Adaptations}
\label{app:baseline-implementations}

{We adapt the baseline implementations to the target dimensions, chronological splits, preprocessing, and one-step-ahead rolling evaluation used in this study. Rolling evaluation advances the input context through the test period; it does not imply that every baseline is retrained at each origin. The target-only neural adapters select settings using validation data, fit before test evaluation, and use the resulting fixed predictors for successive test contexts. Test responses are not used for hyperparameter selection.}

\paragraph{Source use and implementation scope.}
PatchTST and TimeMixer use only the target panel. The AdaRNN adapter uses the upstream recurrent architecture on the target panel. The ROSE adapter uses the upstream Transformer encoder and prediction head for supervised one-step-ahead pre-training across the available source groups, followed by target fine-tuning. It learns the source pre-training checkpoint within our data splits instead of loading an external multi-domain checkpoint. The number of source pre-training and target fine-tuning epochs is selected using target validation loss. This is a local adaptation to our forecasting protocol, rather than a reproduction of the original ROSE pre-training procedure.

DTr-CNN is a local, simplified implementation guided by \citet{ye2021implementing}; an official author-maintained repository was not identified. It selects a source using DTW distances between standardized group-mean series, pre-trains a shared CNN on the selected source, and adapts it to target data. 

\subsection{Runtime Measurement}
\label{app:runtime}

We measure runtime on Stock over 56 test forecasting origins, running one benchmark at a time under an eight-thread CPU budget. 
{Table~\mbox{\ref{tab:runtime-stock-details}} separates fitting from forecasting.}
Hyperparameter and epoch selection, preprocessing, file I/O, startup, and independent prediction verification are excluded.

\emph{Train} includes parameter learning and predictor fitting. For HiCT-VAR, it also includes source-weight generation and predictor refitting at all test origins; it is therefore not solely offline training time. ROSE includes source pre-training and target fine-tuning. The updated AdaRNN measurement includes temporal domain characterization, warm-up, temporal distribution matching, and boosting. \emph{Forecast} measures forward prediction and target-space reconstruction with fitted predictors; it excludes the rolling refits already counted in Train. Total time includes both stages.

\begin{table}[htbp]
	\centering
	\normalsize
	\setlength{\tabcolsep}{4pt}
	\renewcommand{\arraystretch}{1.1}
	\caption{Runtime breakdown on Stock in seconds.}
	\label{tab:runtime-stock-details}
	\begin{tabular}{lrrr}
		\toprule
		Method  & Train & Forecast & Total \\
		\midrule
		PatchTST & 2940.3 & 2.379 & 2942.6 \\
		TimeMixer  & 29.5 & 0.058 & 29.5 \\
		AdaRNN  & 4.3 & 0.027 & 4.3 \\
		DTr-CNN  & 56.9 & 0.020 & 56.9 \\
		ROSE  & 526.8 & 0.718 & 527.5 \\
		HiCT-VAR  & 9.3 & 0.001 & 9.3 \\
		HiCT-RNN  & 10.9 & 0.001 & 10.9 \\
		\bottomrule
	\end{tabular}
\end{table}

\subsection{Validation Sensitivity and Parameter Selection}
\label{app:sensitivity}

\begin{figure}[!htb]
	\centering
	\includegraphics[width=\linewidth]{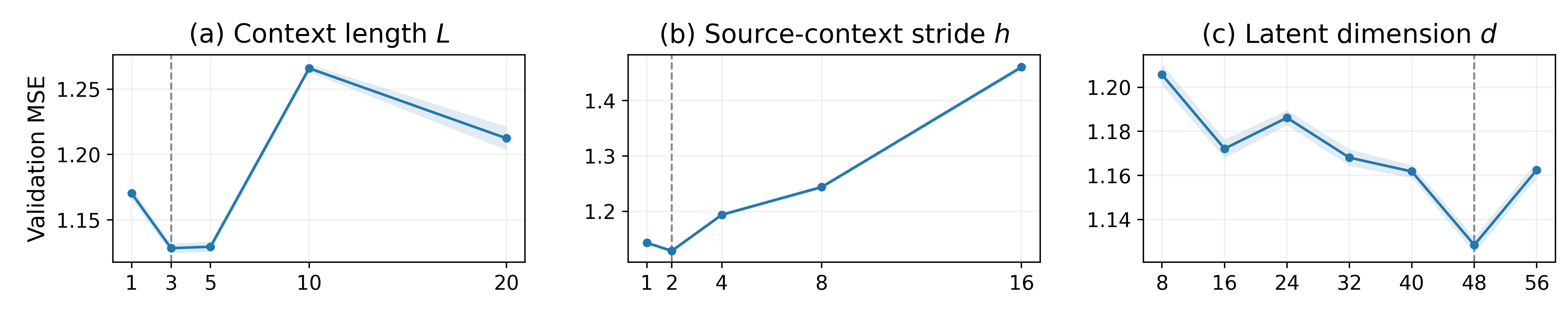}
	\caption{Validation sensitivity on Stock for the same sweeps as
		Figure~\ref{fig:stock-sensitivity}. Curves and shading show the mean
		and one sample standard deviation over five runs; dashed lines
		mark validation-selected values.}
	\label{fig:stock-sensitivity-validation}
\end{figure}

Figure~\ref{fig:stock-sensitivity-validation} reports the validation
curves for the sweeps in Figure~\ref{fig:stock-sensitivity}.
Validation selects $(L,h,d)=(3,2,48)$.
The validation curves favor short contexts and a small source stride.
Validation error rises again at the largest latent dimension.
These selections need not minimize test error: the densest source
sampling gives lower test error despite higher validation error.

\section{Visualization of transferability weights}
\label{app:weight-visualization}

\begin{figure}[ht]
	\centering
	\includegraphics[width=\linewidth]{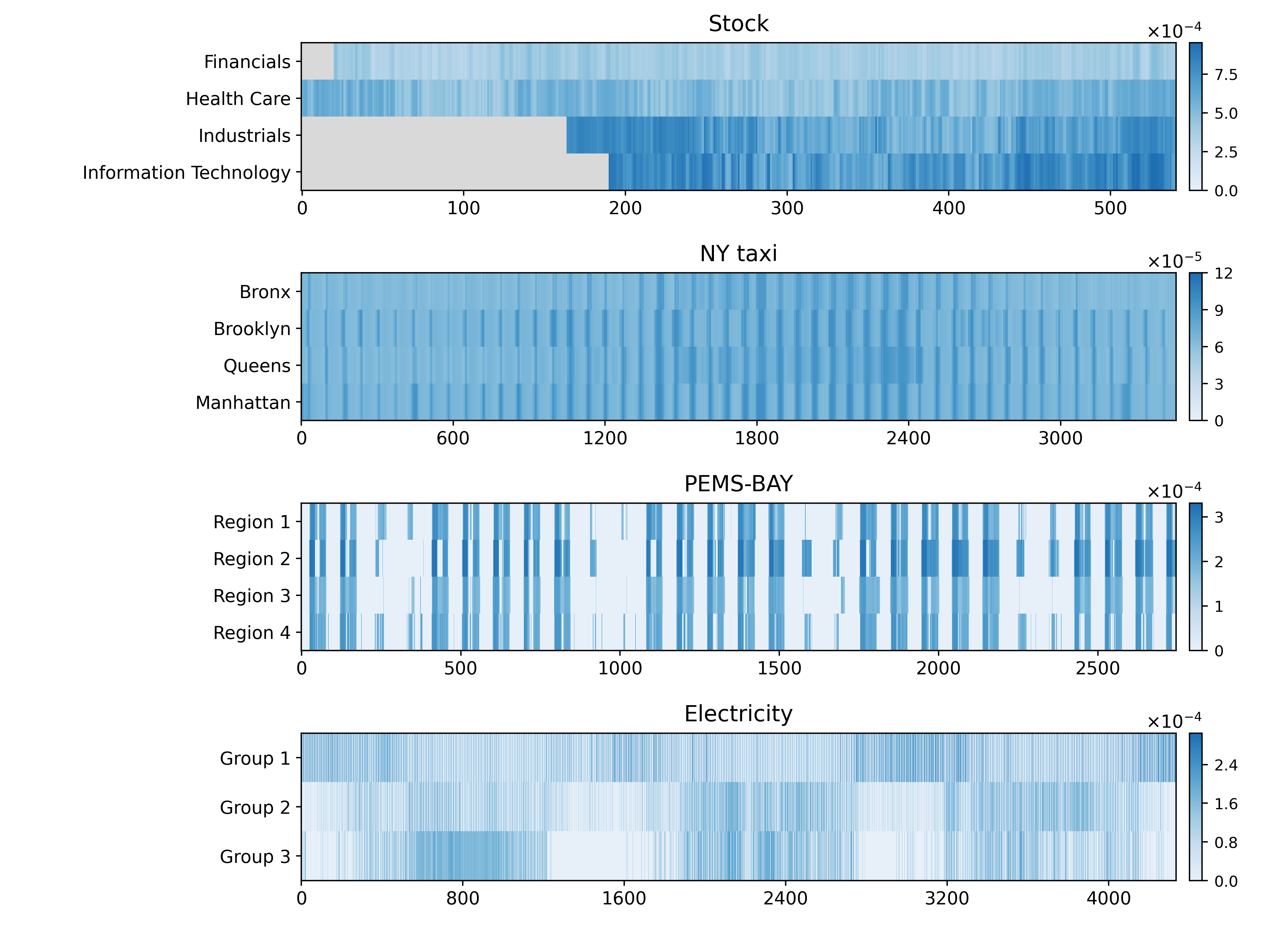}
	\caption{Heatmap visualization of transferability weights for Stock, NY Taxi, PEMS-BAY, and Electricity. Colors show the mean joint weight $\alpha_k\beta_{s\mid k}$ over test forecast origins. Rows represent source groups, and columns index right-aligned source contexts in their original chronological order; gray indicates padding. Each dataset uses its own color scale.}
	\label{fig:heatmap for all}
\end{figure}

Figure~\ref{fig:heatmap for all} summarizes the mean joint weights $\pi_{ks}=\alpha_k\beta_{s\mid k}$ over target test forecasting origins. Rows correspond to source groups, and columns to historical contexts ordered from older to more recent within each group, with the most recent contexts placed at the right edge. Gray cells indicate padding for shorter source histories. These averages summarize source allocation; they do not display its variation from one forecasting origin to the next.

\end{document}